\documentclass[10pt,twocolumn,letterpaper]{article}

\usepackage[pagenumbers]{iccv} 

\usepackage[dvipsnames]{xcolor}
\usepackage{multirow}
\usepackage{tabularx}
\usepackage{tablefootnote}
\usepackage{float}
\usepackage{amsmath}
\usepackage{xr}
\usepackage[accsupp]{axessibility} 
\usepackage{colortbl}
\usepackage{tikz}

\usepackage{xcolor}

\definecolor{iccvblue}{rgb}{0.21,0.49,0.74}
\usepackage[pagebackref,breaklinks,colorlinks,allcolors=iccvblue]{hyperref}

\def\paperID{4253} 
\def\confName{ICCV}
\def\confYear{2025}

\title{
DisenQ: Disentangling Q-Former for Activity-Biometrics
\vspace{-10pt}
}

\author{Shehreen Azad \qquad \qquad Yogesh Singh Rawat \\
Center for Research in Computer Vision\\ University of Central Florida\\
\href{https://sacrcv.github.io/DisenQ-website/}{Project Page} }

\begin{document}
\maketitle
\begin{abstract}
In this work, we address 
activity-biometrics, which involves identifying individuals across diverse set of activities. Unlike traditional person identification, this setting introduces additional challenges as identity cues become entangled with motion dynamics and appearance variations, making biometrics feature learning more complex. While additional visual data like pose and/or silhouette help, they often struggle from extraction inaccuracies. To overcome this, we propose 
a multimodal language-guided framework that replaces reliance on additional visual data with structured textual supervision. At its core, we introduce \textbf{DisenQ} (\textbf{Disen}tangling \textbf{Q}-Former), a unified querying transformer that disentangles biometrics, motion, and non-biometrics features by leveraging structured language guidance. This ensures identity cues remain independent of appearance and motion variations, preventing misidentifications. 
We evaluate our approach on three activity-based video benchmarks, achieving state-of-the-art performance. Additionally, we demonstrate strong generalization to complex real-world scenario with competitive performance on a traditional video-based identification benchmark, showing the effectiveness of our framework.

\end{abstract}
\vspace{-.5cm}    
\section{Introduction}
\label{sec:intro}
Traditional person identification aims to recognize the same individual across different cameras, time and location \cite{ye2021deep}, focusing on facial recognition \cite{adjabi2020past, meng2021magface}, gait analysis \cite{lin2021gait, fan2020gaitpart, fan2023opengait, liang2022gaitedge}, and whole-body biometrics \cite{gu2022clothes, li2023clip, yu2024tf, pathak2020video}. While effective in gait-based analysis, these approaches struggle when individuals engage in diverse daily activities beyond just standing or walking. Extending beyond conventional methods, activity-biometrics involves identifying individuals from their daily activities \cite{azad2024activity} by leveraging motion dynamics, making it more suitable for real-world applications like surveillance, healthcare, smart environments where recognition across diverse activities is crucial.

Effective identification in activity biometrics requires disentangling biometric features from appearance-based non-biometric and motion cues to prevent identity bias and fully leverage motion dynamics. Existing method \cite{azad2024activity} address this using additional visual modalities like silhouettes, but their reliance on accurate extraction limits reliability in real-world settings to aid feature separation. 
This highlights the need for an approach that ensures effective feature separation without additional visual data, improving generalization across varying conditions and activities.

\begin{figure}[t!]
    \centering
    \includegraphics[width=\linewidth]{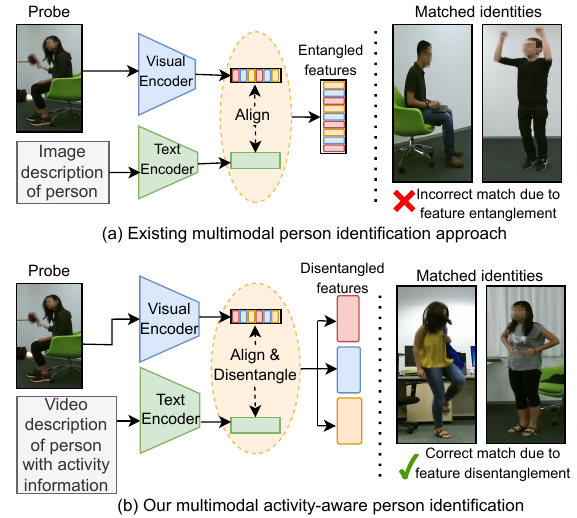}
    \caption{\textbf{Comparison of existing and our activity-aware person identification framework:} (a) Existing multimodal methods \cite{li2023clip} lacking activity-awareness leads to misidentifications due to entangled biometrics and motion features. (b) Our model disentangles biometrics and motion features using language guidance, enabling activity-awareness and more accurate identification across diverse activities while being appearance invariant.
    }
    \vspace{-15pt}
    \label{fig:teaser}
\end{figure}

To reduce reliance on additional visual data, cross-modal learning using text has been explored with CLIP-based contrastive learning methods \cite{li2023clip, yu2024tf, he2024instruct}. 
While effective in aligning global image-text associations, these methods lack identity-specific feature separation, leading to misidentifications due to appearance variation. Additionally, their failure to maintain temporal consistency makes them less suitable for video-based activity-biometrics where motion dynamics play a critical role (Figure \ref{fig:teaser}). 
Multimodal Large Language Models (MLLMs) offer a generative alternative by leveraging structured textual reasoning to achieve better feature separation and robustness in complex scenarios  \cite{li2023blip, dai2023instructblip, liu2023visual,  azad2025hierarq, azad2025understanding, schiappa2024probing, schiappa2024robustness}. However, existing MLLM-based person identification approaches \cite{wang2024large, 2024TVI-LFM} rely on additional visual data and are image based, making them ineffective at capturing motion dynamics essential for activity-biometrics.

To enhance activity-based person identification without relying on additional visual data, we introduce a multimodal \textbf{Disen}tangling \textbf{Q}uerying (\textbf{DisenQ}) Transformer based framework that explicitly separates biometrics, non-biometrics, and motion features. Unlike prior works that rely on additional visual modalities, our approach uses structured text descriptions generated from a frozen Vision-Language Model (VLM) to provide semantic supervision for cross-modal feature learning. These language descriptions serve as explicit guidance, enabling robust feature separation by distinguishing identity-related attributes from appearance and motion cues.

At the core of our approach is our proposed \textbf{DisenQ}, a Disentangling Querying Transformer, that separates biometrics, non-biometrics and motion features using structured textual guidance while minimizing feature leakage. The disentangled features enhance similarity-based retrieval in a traditional identification pipeline. By leveraging language-driven supervision as an auxiliary modality, DisenQ eliminates the need for additional visual modalities, reduces reliance on biased visual cues, and improves generalization across diverse real-world activities.

We evaluate our framework on three activity-based video benchmarks, demonstrating strong generalization across diverse activities. Additionally, to assess its broader applicability, we evaluate it on a large-scale traditional video-based identification benchmark. Our approach achieves state-of-the-art performance on most datasets and remains competitive on others, highlighting its robustness in real-world scenarios. 
Our main contributions are as follows:   
\begin{itemize}
    \item We use language guidance for activity-biometrics to explicitly disentangle feature spaces, enabling cross-modal learning without additional visual data.
    \item We introduce \textbf{DisenQ} (Disentangling Q-Former) that effectively separates biometrics, non-biometrics, and motion features for activity-aware person identification.
    \item Our method achieves state-of-the-art performance on activity-based benchmarks and competitive performance on a traditional video-based person identification benchmark highlighting its generalization ability across diverse real-world scenarios.
\end{itemize}

\section{Related Works}
\label{sec:rel_works}
\textbf{Visual modality based person identification.} Traditional person identification primarily rely on image-based approaches \cite{hong2021fine,huang2019celebrities,yang2019person,cao2022pstr, pathak2020fine}, focusing on body shape, clothing, and appearance. While some works improve robustness with cloth-invariant representations \cite{guo2023SCNet, gu2022clothes, yang2023good}, or additional modalities like silhouettes \cite{jin2022cloth}, skeletons \cite{qian2020long, pathak2025coarse}, or 3D shape \cite{chen2021learning}, they lack spatiotemporal awareness. Video-based approaches \cite{porrello2020robust, jiang2020rethinking, wang2021pyramid, hou2021bicnet, chen2020learning, bai2022salient, he2021dense, liu2023deeply, pathak2020video, Pathak_2025_ICCV} 
integrate temporal cues but remain limited to walking-based identification, making them less effective for activity-based identification. The only prior activity-based identification method \cite{azad2024activity} relies on silhouette extraction, which is unreliable in challenging conditions.  
Instead, we leverage structured language supervision to enhance identity learning without requiring additional visual data.

\noindent
\textbf{Language modality based person identification.} 
Several CLIP based identification approaches \cite{li2023clip, yu2024tf, he2024instruct, chen2023unveiling, yan2023clip, yang2024pedestrian, liang2025differ} leverage image-text contrastive learning \cite{radford2021learningtransferablevisualmodels} for traditional image-based person identification. However, these methods lack temporal modeling and cannot disentangle motion from biometric features, making them unsuitable for activity-based identification. Similarly, large foundation models for image-based person identification \cite{wang2024large, 2024TVI-LFM} incorporate language but rely on additional visual modalities, as well as lack temporal modeling capabilities. To address this, we leverage language guidance not just for cross-modal learning, but to explicitly disentangle video-based features, enabling a more robust and adaptable framework for activity-aware identification.
 
\noindent
\textbf{Multimodal Large Language Models.} Several works  \cite{radford2021learningtransferablevisualmodels, li2024llava, li2023blip, dai2023instructblip, zhu2023minigpt4, alayrac2022flamingo} have been introduced to enhance image-text alignment by bringing visual features closer to language space. Among these, BLIP-2’s Q-Former \cite{li2023blip} is a lightweight approach that effectively aligns visual and language modalities. We build our proposed DisenQ based on this architecture to align visual features with language while disentangling biometrics, non-biometrics, and motion through structured textual guidance. To the best of our knowledge, this is the first work to leverage Q-Former for feature disentanglement.

\section{Method}
\label{sec:method}
\begin{figure*}[t!]
    \centering
    \includegraphics[width=\linewidth]{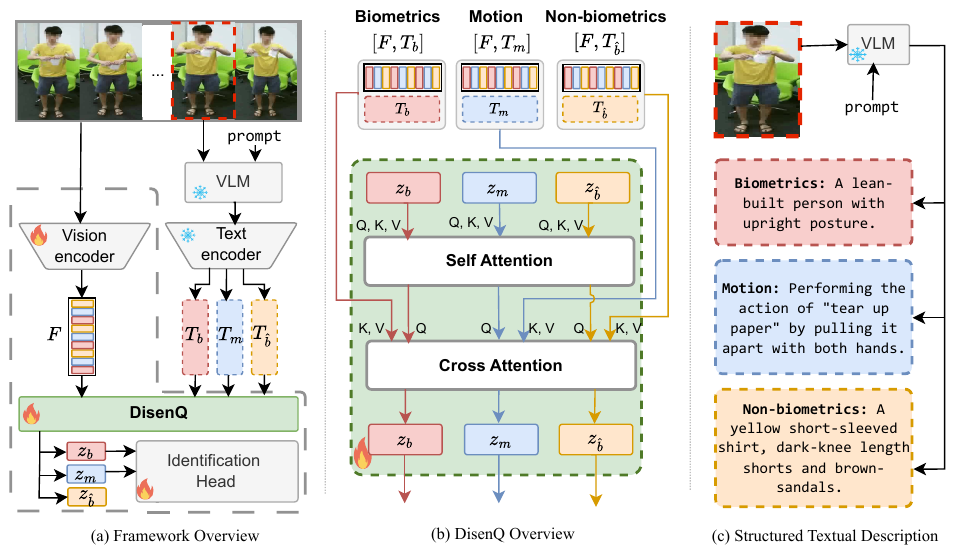}
    \caption{\textbf{Framework overview:} (\textit{a})
    Given an RGB video, our model performs language-guided activity-aware person identification using the proposed \textbf{Disen}tangling \textbf{Q}-Former (DisenQ). \textit{(b)} DisenQ disentangles biometrics, motion, and non-biometrics visual features using structured language guidance and dedicated learnable queries for each feature type. (\textit{c}) The frozen VLM generates structured textual descriptions from the key-frame (red dotted line) to serve as language supervision for DisenQ. Here, $[F, T_x]$ denotes concatenation of $F$ and $T_x$. During inference, only the components within the gray dashed box are utilized, removing the need for text generation.
    }
    \vspace{-10pt}
    \label{fig:architecture}
\end{figure*}
We propose a \textbf{Disen}tangling \textbf{Q}-Former (\textbf{DisenQ}) based multimodal framework for activity-biometrics. 
Our framework consists of a visual encoder to extract visual features from an RGB video and a frozen VLM to generate a detailed prompt which is encoded via a text encoder (Section \ref{subsec:ft_extractor}). To ensure robust feature disentanglement, DisenQ separates biometrics, non-biometrics, and motion features in the visual domain through language guidance (Section \ref{subsec:dis_qf}). Finally, an identification head performs person identification (Section \ref{subsec:reid}), ensuring that biometrics features remain clothing-invariant while incorporating motion cues for improved identity matching. 
Given an RGB video $V$, with ground truth actor and activity labels ($y_{ID}$ and $y_{Action}$), the framework aims to correctly match the identity of the person in a probe video to its corresponding identity in the gallery. The overall approach is illustrated in Figure \ref{fig:architecture}.

\subsection{Feature Extraction}
\label{subsec:ft_extractor}
\textbf{Visual feature extraction.}
Given a sequence of frames from a video $V$, each frame $v_i \in \mathbb{R}^{H \times W \times 3}$, where $H$ and $W$ represent its height and width, is processed through a visual encoder to extract visual features $f_i \in \mathbb{R}^{N \times D}$. Here, $N$ denotes the number of extracted visual tokens per frame, and $D$ is the hidden dimension of each token. Here, each visual feature $f_i$ has temporal ordering information associated with it through a position embedding layer. Finally, temporal attention pooling is applied on all frame features to get a global video-level feature $F$.

\noindent
\textbf{Prompt generation and textual feature extraction.}
To generate structured and semantically consistent language description, we use a frozen VLM to generate prompts from the key-frame of the input video during only training, without requiring the VLM during inference. 
These descriptions are categorized into three distinct components following pre-defined templates: Biometrics prompt ($P_{b}$), describing identity-specific traits such as body shape, posture, and notable physical characteristics;  Motion prompt ($P_{m}$) describing the action label and movement; and Non-biometrics prompt ($P_{\hat{b}}$), describing clothing, and accessories. To maintain consistency, biometrics descriptions are generated only once per unique identity and reused in all subsequent videos of the same actor by storing and iteratively refining it by updating the stored description using a running average.
This prevents major description drift and ensures stable identity representation across varied activities and appearances. 

The generated prompts are then encoded using a pre-trained frozen text-encoder to obtain textual embeddings ($T_{b}, T_{m}, T_{\hat{b}}$) which serve as language-driven supervision for visual feature disentanglement. 

\subsection{DisenQ}
\label{subsec:dis_qf}
We introduce \textbf{DisenQ} (\textbf{Disen}tangling \textbf{Q}uerying Transformer) to separate biometrics, motion and non-biometrics features in the visual domain by aligning visual representations with structured textual cues.
Adapted from the original Q-Former \cite{li2023blip}, DisenQ introduces three separate sets of learnable queries: $z_b$ (biometrics), $z_m$ (motion) and $z_{\hat{b}}$ (non-biometrics); instead of a single query set, enabling explicit disentanglement. Each query set shares the same self-attention and cross-attention layers while leveraging textual guidance, ensuring effective feature separation. However, they explicitly attend to different information without interaction, preserving distinct feature representations for biometrics, motion and non-biometrics.
The learned queries are then utilized for activity-based person identification, improving the model’s ability to distinguish individuals based on biometrics while leveraging motion cues and remaining invariant to non-biometrics attributes.

\noindent
\textbf{Biometrics feature disentanglement.}
To extract identity-related features, the biometrics query $z_{b}$ attends to itself through self-attention to refine itself.
Then the refined query performs cross-attention with the visual feature $F$ and biometrics textual supervision features $T_{b}$ with query, key and value being used as Equation \ref{eq:bio_qkv}.  
\begin{equation}
    Q_b = W z_b, \mkern10mu K_b = W [F, T_b], \mkern10mu V_b = W [F, T_b].
    \label{eq:bio_qkv}
\end{equation}
Here, $[F, T_{b}]$ denotes concatenation of $F$ and $T_{b}$, followed by a linear projection. 

\noindent
\textbf{Motion feature disentanglement.}
To extract motion-specific representations, the motion query $z_{m}$, similar to biometrics query $z_{b}$, first undergoes self-attention, ensuring it refines motion-related patterns independently.
Subsequently, the motion query cross-attends to the visual feature $F$ and its corresponding textual feature $T_{m}$ with query, key and value acting as Equation \ref{eq:mo_qkv}. 
\begin{equation}
    Q_{m} = W z_{m}, \mkern8mu K_{m} = W [F, T_{m}], \mkern8mu V_{m} = W [F, T_{m}].
    \label{eq:mo_qkv}
\end{equation}

\noindent
\textbf{Non-biometrics feature disentanglement.}
To separate non-biometrics features, the non-biometrics query $z_{\hat{b}}$ similar to others, also, first undergoes self-attention, refining itself without influence from other feature categories.
Following this, the non-biometrics queries cross-attend to the visual feature $F$ and non-biometrics textual feature $T_{\hat{b}}$ with query, key and value acting as Equation \ref{eq:nbio_qkv}. 
\begin{equation}
    Q_{\hat{b}} = W z_{\hat{b}}, \mkern10mu K_{\hat{b}} = W [F, T_{\hat{b}}], \mkern10mu V_{\hat{b}} = W [F, T_{\hat{b}}].
    \label{eq:nbio_qkv}
\end{equation}

\subsection{Identification Head}
\label{subsec:reid}
The learned query embeddings $z_{b}$, $z_{m}$ and $z_{\hat{b}}$ go through mean pooling to form single vectors, denoted as $F_{b}$, $F_{\hat{b}}$, and $F_{m}$, among which only $F_{b}$ and $F_{m}$ is used for final identification.

\noindent
\textbf{Loss Functions.}
During training, the model is optimized to refine $F_b$ using a combination of standard cross-entropy ($\mathcal{L}_{ID}$), and triplet loss ($\mathcal{L}_{Tri}$) following \cite{azad2024activity, gu2022clothes}. These losses are defined as Equation \ref{eq:ce} and Equation \ref{eq:Tri}.

\begin{equation}
    \mathcal{L}_{ID} = -y \log \hat{y} \simeq \mathcal{L}_{Act},
    \vspace{-3pt}
    \label{eq:ce}
\end{equation}
\vspace{-10pt}
\begin{equation}
    \mathcal{L}_{Tri} = \max \left( \mathcal{D}(F_b^a, F_b^p) - \mathcal{D}(F_b^a, F_b^n) + m, 0 \right),
    \label{eq:Tri}
\end{equation}

\noindent
Here, $y$ and $\hat{y}$ denote the ground truth and predicted labels. $F_b^p$ and $F_b^n$ represent the positive and negative biometrics features for an anchor biometrics feature $F_b^a$ within the same batch. $\mathcal{D}(\cdot)$ computes the Euclidean distance, and $m$ is the margin in the triplet loss.

Since the motion feature $F_m$ contributes to identity recognition, it is explicitly trained to preserve motion-related information while remaining independent of biometrics attributes. The model is optimized for $F_m$ using the cross-entropy loss ($\mathcal{L}_{Act}$) of Equation~\ref{eq:ce}. 

Furthermore, to reinforce the independence of biometrics and non-biometrics features, an orthogonality constraint is imposed between $F_b$ and $F_{\hat{b}}$ as Equation \ref{eq:Orth}.
\begin{equation}
    \mathcal{L}_{Orth} = \left\| F_b^T F_{\hat{b}} \right\|.
    \label{eq:Orth}
\end{equation}

\noindent
The overall loss function is defined as Equation \ref{eq:total_loss}.
\begin{equation}
    \mathcal{L} = \lambda_1 \mathcal{L}_{ID} + \lambda_2 \mathcal{L}_{Tri} + \lambda_3 \mathcal{L}_{Orth} + \lambda_4 \mathcal{L}_{Act} . 
    \label{eq:total_loss}
\end{equation}

\noindent
Here, $\lambda_{i \in {1,..,4}}$, is weighting factor for each loss term.

\noindent
\textbf{Identity Similarity Computation.}
\noindent
To enhance identity matching, we introduce an adaptive weighting mechanism that integrates motion features into the similarity calculation, unlike traditional methods that rely solely on biometrics. Instead of fixed weights, we use a lightweight MLP to dynamically adjust the contribution of biometrics and motion features based on their relevance. Given a probe identity $A$ and gallery identity $B$, we compute cosine similarities for both biometrics and motion features, concatenate them, and pass them through the MLP with ReLU activations and a softmax function. This enables the model to leverage motion cues to guide biometrics matching, prioritizing motion when it provides meaningful identity information and relying more on biometrics when motion cues are less discriminative.
The final similarity score is computed as Equation \ref{eq:simfinal}. 

\begin{equation}
    Sim(A,B) = \alpha_1 Sim_{b}(A,B) + \alpha_2 Sim_{m}(A,B).
    \vspace{-4pt}
    \label{eq:simfinal}
\end{equation}

\noindent
Here, $\alpha_{i \in 1,2}$ are the weighting factors. 

\noindent
\textbf{Inference.}
DisenQ operates without textual supervision during inference, relying solely on the learned query embeddings acquired during training. It utilizes self-attention to retain query-specific information and cross-attention to extract relevant visual features, ensuring effective disentanglement of biometrics, non-biometrics and activity features purely from visual embeddings.

\section{Experiments}
\label{sec:experiments}
\begin{table*}[t!]
    \centering
    \small
    \caption{\textbf{Performance comparison of activity-based person identification} on NTU RGB-AB, PKU MMD-AB, Charades-AB on same activity (denoted by Same) and cross activity (denoted by Cross) evaluation protocol. R@1 denotes Rank 1 accuracy. $\dag$ denotes results produced in our environment. \textbf{Bold} and \underline{underline} denotes best and second best results.
    }
    \resizebox{1.01\linewidth}{!}{
    \begin{tabular}{ll|cccc|cccc|cccc}
        \hline
        \multirow{3}{*}{Methods} & \multirow{3}{*}{Venue} & \multicolumn{4}{c|}{NTU RGB-AB} & \multicolumn{4}{c|}{PKU MMD-AB} &   \multicolumn{4}{c}{Charades-AB}  \\
        && \multicolumn{2}{c}{Same} & \multicolumn{2}{c|}{Cross} & \multicolumn{2}{c}{Same} & \multicolumn{2}{c|}{Cross}& \multicolumn{2}{c}{Same} & \multicolumn{2}{c}{Cross}\\
        &&R@1 & mAP & R@1 & mAP & R@1 & mAP &R@1 & mAP & R@1 & mAP & R@1 & mAP  \\ 
        \hline
        \multicolumn{8}{l}{\textit{ \textcolor{gray}{Models with only visual modality}}}\\
        TSF \cite{jiang2020rethinking} & AAAI 20&  $71.8$ & $31.8$ &$67.8$&$26.9$& $76.4$ & $37.5$ &$71.6$&$33.2$& $35.4$ & $21.9$ &$30.2$ & $19.0$\\
        VKD \cite{porrello2020robust} & ECCV 20& $67.4$ & $35.6$ &$66.3$&$31.5$& $78.4$ & $38.5$ &$72.2$&$34.3$& $36.3$ & $20.7$ &$31.9$ & $18.8$\\ 
        BiCnet-TKS \cite{hou2021bicnet} & CVPR 21& $72.7$ & $34.5$ &$69.1$&$30.2$& $80.8$ & $38.5$ &$77.1$&$33.3$& $40.3$ & $27.3$ &$38.3$ & $23.3$\\
        PSTA \cite{wang2021pyramid} & ICCV 21& $67.4$ & $34.8$ &$65.1$&$31.4$& $77.4$ & ${50.4}$ &$72.4$&$47.4$& $42.9$ & $28.3$ &$38.7$ & $24.8$\\
        STMN \cite{eom2021video} & ICCV 21& $73.0$ & $35.1$ &$70.2$&$30.1$& $76.6$ & $47.9$ &$71.5$&$42.2$& $38.7$ & $24.5$ &$33.9$ & $20.8$\\
        SINet \cite{bai2022salient} & CVPR 22& $69.4$ & $30.7$ &$66.2$&$27.8$& $79.6$ & $40.8$ &$74.1$&$26.2$& $40.3$ & $26.9$ & $37.3$ & $21.9$ \\
        CAL \cite{gu2022clothes} & CVPR 22& $73.8$ & $28.4$ &$70.3$&$24.0$& $81.3$ & $49.4$ &$78.3$&$43.4$& $43.8$ & $25.8$ &$40.1$&$21.2$\\
        Video-CAL \cite{gu2022clothes} & CVPR 22& ${75.5}$ & ${39.9}$ &$73.3$&$31.7$& $79.6$ & $49.4$ &$77.3$&$45.7$& ${43.9}$ & ${28.5}$ & $41.5$ & $25.8$ \\
        PSTR \cite{cao2022pstr} & CVPR 22& $69.1$ & $34.1$ &$68.3$&$32.5$& $84.3$ & $47.5$ &$78.0$&$41.2$& $37.2$ & $24.7$ & $35.1$ & $20.3$\\
        AIM \cite{yang2023good} & CVPR 23& $71.4$ & $35.4$ &$72.8$&$30.2$& ${82.5}$ & $48.9$ &$79.2$&$44.9$& $40.1$ & $28.3$ &$35.6$ & $26.7$\\
        SCNet \cite{guo2023SCNet} & ACM MM 23& $69.9$ & $31.5$ &$68.8$&$26.3$& $79.5$ & $43.6$ &$73.9$&$39.7$& $31.7$ & $21.9$ &$27.4$ & $17.6$\\
        ABNet \cite{azad2024activity} & CVPR 24& $\underline{78.8}$ & $40.3$ &$\underline{77.0}$&$\underline{37.6}$& $\underline{86.8}$ & $\underline{57.3}$ &$81.4$&$51.8$& $\underline{45.8}$ & $\underline{31.6}$ & $\underline{44.8}$ & $\underline{28.8}$ \\
        \hline
        \hline
        \multicolumn{7}{l}{\textit{ \textcolor{gray}{Models with visual + language modality}}}\\
        CLIP ReID \cite{li2023clip} \dag & AAAI 23& $77.1$ & $40.2$ & $75.2$ & $33.7$ & $82.3$ & $52.1$ & $81.2$ & $50.8$ & $44.2$ & $31.3$ & $42.1$ & $27.7$  \\
        CCLNet \cite{chen2023unveiling} \dag & ACM MM 23& $75.2$ & $36.1$ & $74.3$ & $33.1$ & $83.2$ & $51.4$ & $80.1$ & $47.5$ & $42.1$ & $29.3$ & $38.8$ & $23.4$ \\
        TF-CLIP \cite{yu2024tf} \dag & AAAI 24& $77.3$ & $41.2$ & $74.8$ &$31.3$ & $83.4$ & $52.3$ & $80.8$ & $50.1$ & $40.2$ & $28.1$ & $39.7$ & $26.0$  \\
        TVI-LFM \cite{2024TVI-LFM} \dag & NeurIPS 24& $76.2$ & $38.1$ & $75.9$ & $34.1$ & $85.2$ & $53.9$ & $81.5$ & $52.1$ & $45.7$ & $30.1$ & $42.8$ & $28.3$ \\
        Instruct-ReID \cite{he2024instruct} \dag & CVPR 24&  $78.2$ & $\underline{41.5}$ & $75.9$ & $33.4$ & $84.3$ & $53.1$ & $\underline{81.7}$ & $\underline{52.3}$ & $44.8$ & $28.3$ & $40.1$& $25.3$ \\
        \hline  
        EVA-CLIP \cite{sun2023eva} \dag &  & $71.2$ & $35.1$ &$69.1$&$28.3$& $73.8$ & $46.2$ &$67.4$&$39.4$& $38.1$ & $26.1$ &$31.3$ &$21.8$\\
        \rowcolor[gray]{.92}
        Ours && $\mathbf{82.2}$ & $\mathbf{43.8}$ &$\mathbf{80.9}$ & $\mathbf{41.3}$& $\mathbf{89.2}$ & $\mathbf{59.3}$ &$\mathbf{84.1}$ &$\mathbf{56.9}$ & $\mathbf{49.9}$ & $\mathbf{34.8}$ & $\mathbf{48.4}$ & $\mathbf{32.5}$\\
        \hline
    \end{tabular}
    }
    \label{tab:sota}
    \end{table*}

\noindent
\textbf{Datasets:}
\label{subsec:datasets}
We evaluate our model on NTU RGB-AB, PKU MMD-AB, and Charades-AB, following \cite{azad2024activity}. NTU RGB-AB consists of $106$ actors performing $94$ actions across $88.7$k samples, while PKU MMD-AB includes $66$ actors, $41$ actions, and $17$k samples. Charades-AB features $267$ actors with $157$ actions across $9.8$k videos, averaging $6.8$ activities per video. 
To assess the generalization capability of our model on more challenging real-world scenarios, we evaluate it on MEVID \cite{davila2023mevid}, which includes $158$ actors and $8$k tracklets, incorporating greater viewpoint, distance, and lighting variations, making it a more complex benchmark for video-based identification. 

\noindent
\textbf{Evaluation Protocol and Metrics:}
We follow the same evaluation protocol and dataset splits as \cite{azad2024activity} for NTU RGB-AB, PKU MMD-AB, and Charades-AB, employing two evaluation protocols: same-activity and cross-activity. Additionally, due to view information explicitly being available for NTU RGB-AB and PKU MMD-AB, we evaluate including and excluding same-view settings too. For MEVID, we use the official protocol and splits. We report rank 1, rank 5 accuracies and mAP as evaluation metrics.

\subsection{Implementation Details}
\label{subsec:impl}
We use $8$ frames which are randomly selected with a stride of $4$ from each original video to create RGB clip. Each frame is resized to $224 \times 224$ and horizontal flipping is used for data augmentation, following \cite{gu2020appearance, azad2024activity}. We use pre-trained ViT G/14 from EVA-CLIP \cite{sun2023eva} as the visual encoder and BERT \cite{devlin2018bert} as the frozen text encoder. Additionally, we use LLaVA 1.5 7B \cite{liu2023visual} as the frozen VLM to generate prompts. We initialize DisenQ with pre-trained weights from InstructBLIP \cite{dai2023instructblip}. We train the model for $60$ epochs with a batch size of $32$ with each batch containing $8$ person and $4$ clips for each person. AdamW is used as the optimizer with weight decay of $5e-2$ and base learning rate of $1e-4$ with $\beta$ values as $[0.9, 0.999]$.
The triplet loss margin $m$ is set to $0.3$, $\lambda_i \in [1,..,4]$ in Equation \ref{eq:total_loss} is set as $0.01$.

\subsection{Results}
\label{subsec:results}
\textbf{Performance on activity-biometrics benchmarks.}
Table \ref{tab:sota} presents the performance comparison of our framework against other existing methods. Across all datasets, our model outperforms the previous best-performing approach, improving Rank-1 accuracy and mAP across all evaluation protocols on NTU RGB-AB, PKU MMD-AB, and Charades-AB. Notably, we observe an average Rank-1 accuracy improvement of $3.7$\%, $2.4$\% and $3.9$\% respectively on NTU RGB-AB, PKU MMD-AB, and Charades-AB, demonstrating the effectiveness of our approach. We present more results in supplementary.

\noindent
\textbf{Generalization to traditional video-based benchmark.}
Table \ref{tab:mevid} presents the identification results of our model compared to concurrent methods on MEVID, a large-scale traditional video-based identification dataset primarily focused on walking sequences.  Unlike NTU RGB-AB, PKU MMD-AB, and Charades-AB, which contain diverse activities, MEVID lacks activity variability, making activity-based identification less impactful. Despite this, our model remains competitive, achieving a $1.2\%$ improvement in Rank-1 accuracy. This demonstrates that while our framework is designed for activity-biometrics, it generalizes well to traditional video-based identification scenarios by effectively disentangling identity from appearance, ensuring robust performance even in real-world unconstrained  settings.

   \begin{table}[t!]
        \centering
        \small
        \caption{\textbf{Performance comparison of traditional person identification} on MEVID in general evaluation setting. R@1 and R@5 denote rank 1 and rank 5 accuracies. $\dag$ denotes reproduced results. 
        }
        \resizebox{\linewidth}{!}{
    \begin{tabular}{ll|ccc}
        \hline
        Methods & Venue& R@1 & R@5 & mAP \\
        \hline
        \multicolumn{4}{l}{\textit{  \textcolor{gray}{Models with only visual modality}}}\\
        Attn-CL \cite{pathak2020video} &AAAI 20& $42.1$ & $56.0$ & $18.6$\\
        Attn-CL + rerank \cite{pathak2020video} &AAAI 20& $46.5$ & $59.8$ & $25.9$\\
        AP3D \cite{gu2020appearance} & ECCV 20& $39.0$ & $56.0$ &$15.9$  \\
        TCLNet \cite{hou2020temporal} & ECCV 20& $48.1$ & $60.1$ &$23.0$\\
        BiCnet-TKS \cite{hou2021bicnet} & CVPR 21& $19.0$ & $35.1$& $6.3$ \\
        STMN \cite{eom2019learning} & ICCV 21& $31.0$ & $54.4$&$11.3$ \\
        PSTA \cite{wang2021pyramid} & ICCV 21&  $46.9$ & $60.8$& $21.2$\\
        PiT \cite{zang2022multidirection} & TII 22& $34.2$  &$55.4$& $13.6$ \\
        CAL \cite{gu2022clothes} & CVPR 23& $52.5$ & $66.5$&$27.1$ \\
        ShARc \cite{zhu2024sharc} & WACV 24& $\underline{59.5}$ & $\mathbf{70.3}$ & $29.6$ \\
        ABNet \cite{azad2024activity} \dag & CVPR 24& $58.3$ & $68.4$&$\underline{30.1}$ \\
        \hline
        \hline
        \multicolumn{3}{l}{\textit{ \textcolor{gray}{Models with visual + language modality}}}\\
        CLIP ReID \cite{li2023clip} \dag & AAAI 23&  $51.2$ & $64.2$& $28.3$\\
        CCLNet \cite{chen2023unveiling} \dag & ACM MM 23&  $50.8$ & $60.3$ & $27.1$ \\
        TVI-LFM \cite{2024TVI-LFM} \dag & NeurIPS 24& $49.2$ & $61.8$&$23.7$\\
        Instruct-ReID \cite{he2024instruct} \dag & CVPR 24&  $53.8$ &$59.4$& $28.4$\\
        \hline
        EVA-CLIP \cite{sun2023eva} \dag &  & $53.1$ & $59.2$&$26.9$\\
        \rowcolor[gray]{.92}
        Ours &  & $\mathbf{60.7}$ & $\mathbf{70.3}$ & $\mathbf{30.4}$\\
        \hline
        \end{tabular}
        }
        \label{tab:mevid}
\end{table}
    
\subsection{Ablation Studies}
\label{subsec:ablation}
We conduct ablation studies on NTU RGB-AB and Charades-AB datasets on the same activity, including same view evaluation protocol and present the results in Table \ref{tab:abla_main}. While NTU RGB-AB provides a controlled setting with diverse clothing and activity variations; Charades-AB contains much more real-world complexity, including varied lighting, occlusions, and higher appearance variations, which better tests model generalization.

\noindent
\textbf{Contribution of each component} is presented in Table \ref{tab:abla_main} \textit{(top)}. A vision encoder alone struggles due to entangled identity, appearance, and motion features leading to poor performance. Introducing text supervision via cross-attention and projecting features into distinct spaces improves identity retention by mitigating the influence of appearance variability. However, the most substantial gains come from DisenQ, which explicitly separates biometrics, non-biometrics, and motion features. By aligning separate learnable queries with structured textual priors, DisenQ establishes a well-structured feature representation that significantly enhances activity-biometrics performance. 

\noindent
\textbf{Ablation of different type of feature disentanglement}, illustrated in Table \ref{tab:abla_main} (\textit{middle}) presents their individual impact on performance.  
When biometrics and non-biometrics features are disentangled, the model effectively mitigates clothing bias but struggles with variations in motion, resulting in improved yet suboptimal performance across different actions. Disentangling biometrics and motion features enhances stability by preserving identity-specific movement patterns, which are crucial for reliable identification across activities. The most comprehensive performance is achieved when all three feature types are disentangled, ensuring that identity-related features remain distinct while controlling appearance and motion influences. 

\begin{table}[t!]
    \caption{\textbf{Ablation studies for each component.} Here, $F_{b}$, $F_{\hat{b}}$ and $F_{m}$ denote biometrics, non-biometrics and motion features.}
    \begin{tabular}{l|cc|cc}
         \hline
         \multirow{2}{.5cm}{Method}& \multicolumn{2}{c|}{NTU RGB-AB} & \multicolumn{2}{c}{Charades-AB}\\
         &Rank 1 & mAP & Rank 1 & mAP\\
         \hline
         \multicolumn{5}{l}{\textit{   \textcolor{gray}{Contribution of each component}}}\\
         Vision encoder & $73.2$ & $36.2$ & $40.1$ & $29.2$\\
         + Text encoder & $77.7$ & $40.6$ & $46.5$ & $31.8$\\
         \hline
         \rowcolor[gray]{.92}
         + DisenQ & $\mathbf{82.2}$ & $\mathbf{43.8}$ & $\mathbf{49.9}$ & $\mathbf{34.8}$\\
         \hline
         \hline
         \multicolumn{5}{l}{\textit{   \textcolor{gray}{Ablation of different type of feature disentanglement}}}\\
         No disentanglement & $74.2$ & $38.2$ & $42.3$ & $29.9$ \\
         $F_{b}$ and $F_{\hat{b}}$ & $76.6$ & $40.9$ & $44.7$ & $31.9$\\
         $F_{b}$ and $F_m$ & $79.2$ & $41.1$ & $48.2$ & $32.9$ \\
         \hline
        \rowcolor[gray]{.92}
        $F_{b}$, $F_{\hat{b}}$ and $F_{m}$ & $\mathbf{82.2}$ & $\mathbf{43.8}$ & $\mathbf{49.9}$ & $\mathbf{34.8}$\\
         \hline
         \hline
         \multicolumn{5}{l}{\textit{   \textcolor{gray}{Performance of each disentangled feature}}}\\
         Biometrics & $80.4$ & $43.0$ & $48.1$ & $32.0$\\
         Non-biometrics & $3.8$ & $1.2$ & $1.3$ & $0.1$ \\
         Motion & $76.3$ & $39.4$ & $44.2$ & $27.1$\\
         \hline
         \rowcolor[gray]{.92}
         Biometrics + Motion & $\mathbf{82.2}$ & $\mathbf{43.8}$ & $\mathbf{49.9}$ & $\mathbf{34.8}$ \\
         \hline
         
    \end{tabular}
    \vspace{-16pt}
    \label{tab:abla_main}
\end{table}

\noindent
\textbf{Individual performance of each disentangled feature}, illustrated in Table \ref{tab:abla_main} (\textit{bottom}), provides further insights into their discriminative power for activity-based identification. Biometrics features alone exhibit the highest performance among each individual feature types, highlighting their intrinsic value in accurately identifying individuals. In contrast, non-biometric features significantly degrade performance, indicating that our disentanglement was effective in removing identity-related information from this feature space. Motion features offer moderate performance, providing additional context but lacking the distinctiveness of biometrics attributes. The synergy between biometric and motion features yields the most effective results, leveraging both identity cues and dynamic movement patterns for robust identification across challenging scenarios.

\section{Analysis and Discussion}
\label{sec:analysis}

\begin{figure*}[t!]
    \centering
    \includegraphics[width=\linewidth]{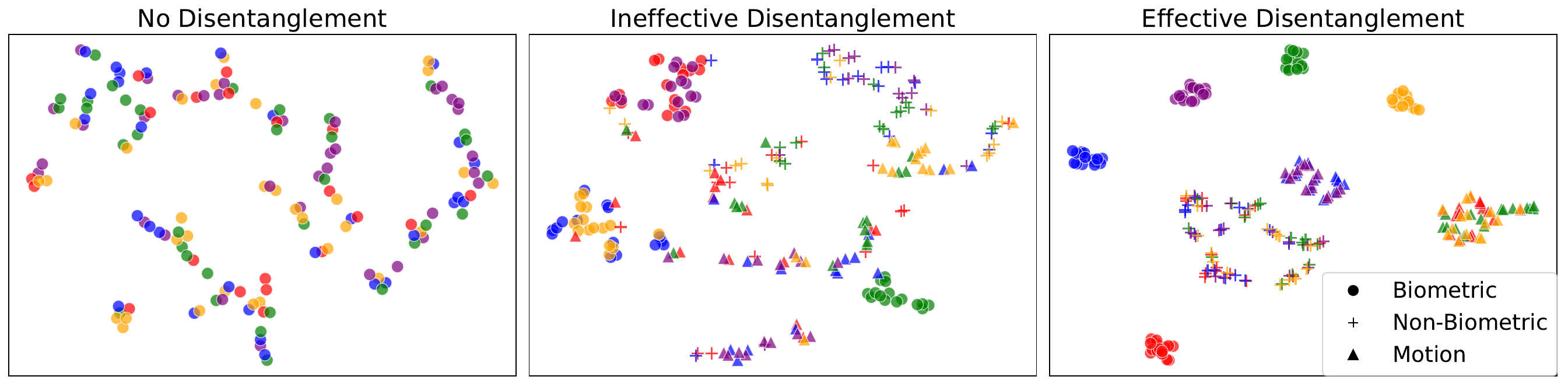}
    \caption{\textbf{Impact of DisenQ on feature disentanglement.} The left plot shows the feature space of vision encoder with no feature disentanglement, resulting in poor identity clustering. The middle plot shows an ineffective disentanglement using cross-attention and projection, where biometrics features remain mixed with non-biometrics, causing improper clustering. In contrast, the right plot demonstrates DisenQ-enabled disentanglement, achieving well-separated biometrics clusters, distinctly isolated from non-biometrics and motion features. Colors represent five identities across two activity classes.
    }
    \vspace{-10pt}
    \label{fig:tsne}
\end{figure*}
\begin{figure*}[t!]
    \centering
    \includegraphics[width=0.9\linewidth]{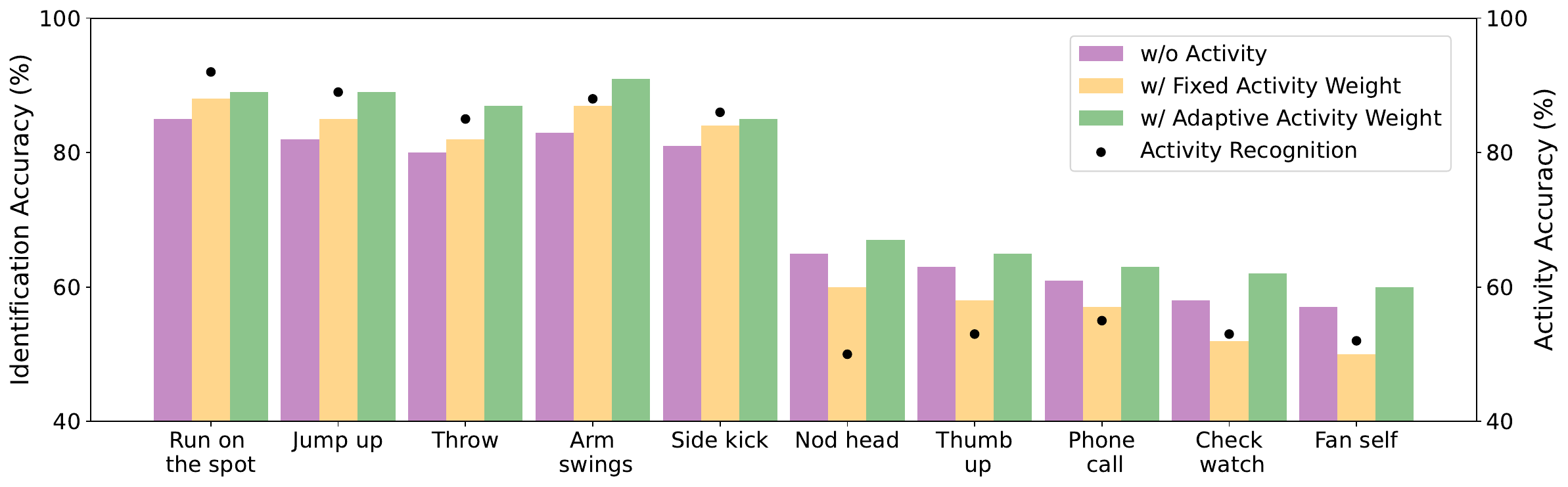}
    \caption{\textbf{Performance analysis across activities} (top 5 best and worst) on NTU RGB-AB. 
    Here, bars and dots respectively represent person identification and action recognition accuracy.
    }
    \label{fig:activity}
    \vspace{-10pt}
\end{figure*}

\textbf{Effect of disentanglement on feature space.} Figure \ref{fig:tsne} presents the impact of DisenQ on feature disentanglement by comparing feature spaces before and after its application. Without DisenQ, feature separation relies on simple cross-attention and projection, resulting in overlapping clusters where biometric features mix with non-biometrics from different identities wearing similar clothing. In contrast, DisenQ effectively separates biometrics, non-biometrics, and motion features into distinct spaces with clear boundaries. DisenQ's structured separation ensures that biometrics clusters remain identity-specific and free from confounding appearance-based cues, leading to robust identification.

\noindent
\textbf{Impact of design choice for disentanglement.} We explore different architectural variations of DisenQ to evaluate the trade-off between complexity and effectiveness. A variant using three independent Q-Formers—each learning biometrics, non-biometrics, or motion features separately—yields only a marginal $0.23$\% Rank-1 accuracy gain on NTU RGB-AB while tripling the parameter count, suggesting that our original design is already sufficient for disentanglement. To test whether additional parameters could still be beneficial, a deeper DisenQ variant with the same parameter count as the three-Q-Former setup results in a $3.8$\% drop due to overfitting, indicating that simply increasing model capacity does not guarantee better feature separation. These findings highlight that structured learning is more critical than model size, and our DisenQ architecture strikes an optimal balance between effectiveness, and computational cost for activity-based person identification.

\noindent
\textbf{Performance analysis across activities.} To examine the impact of different activities on person identification, we analyze performance across activity classes by identifying the five best and worst-performing actions. While activities involving significant body movements (e.g., running, jumping) provide distinctive motion patterns that aid recognition, they can introduce biases if overemphasized. Conversely, subtle activities (e.g.,  minor hand/head gestures) may lower accuracy due to weaker motion cues. Our findings (Figure \ref{fig:activity}) show that fixed weighting ($\alpha_1 = \alpha_2 = 0.5$ in Equation \ref{eq:simfinal}) of biometric and motion features can negatively affect identification for low-motion activities, whereas adaptive weighting ensures motion features contribute only when beneficial, stabilizing performance. Notably, highly distinctive actions retain high person identification accuracy even without explicit motion cues, confirming that motion serves as a complementary rather than dominant factor. Likewise, challenging activities do not inherently degrade identification performance, as the model prioritizes biometrics features when necessary, ensuring balanced identification.

\begin{figure}[t!]
    \begin{minipage}{.2\textwidth}
        \captionof{table}{\textbf{Similarity of generated prompts} across multiple runs on NTU RGB-AB subset. Sim. and Std. Dev. denotes mean cosine similarity and standard deviation.}
        \resizebox{\linewidth}{!}{
        \begin{tabular}{p{.4cm}|cc}
             \hline 
             Ft. & Sim. & St. Dev.  \\
             \hline
             $T_b$ & $0.92$ & $0.03$\\
             $T_{\hat{b}}$ & $0.79$ & $0.12$\\
             $T_m$ & $0.68$ & $0.17$\\
             \hline
        \end{tabular}
        }
        \label{tab:runs}
    \end{minipage}
    \vspace{-4pt}
    \hfill
    \begin{minipage}{.25\textwidth}
            \includegraphics[width=.99\linewidth, height=3.1cm]{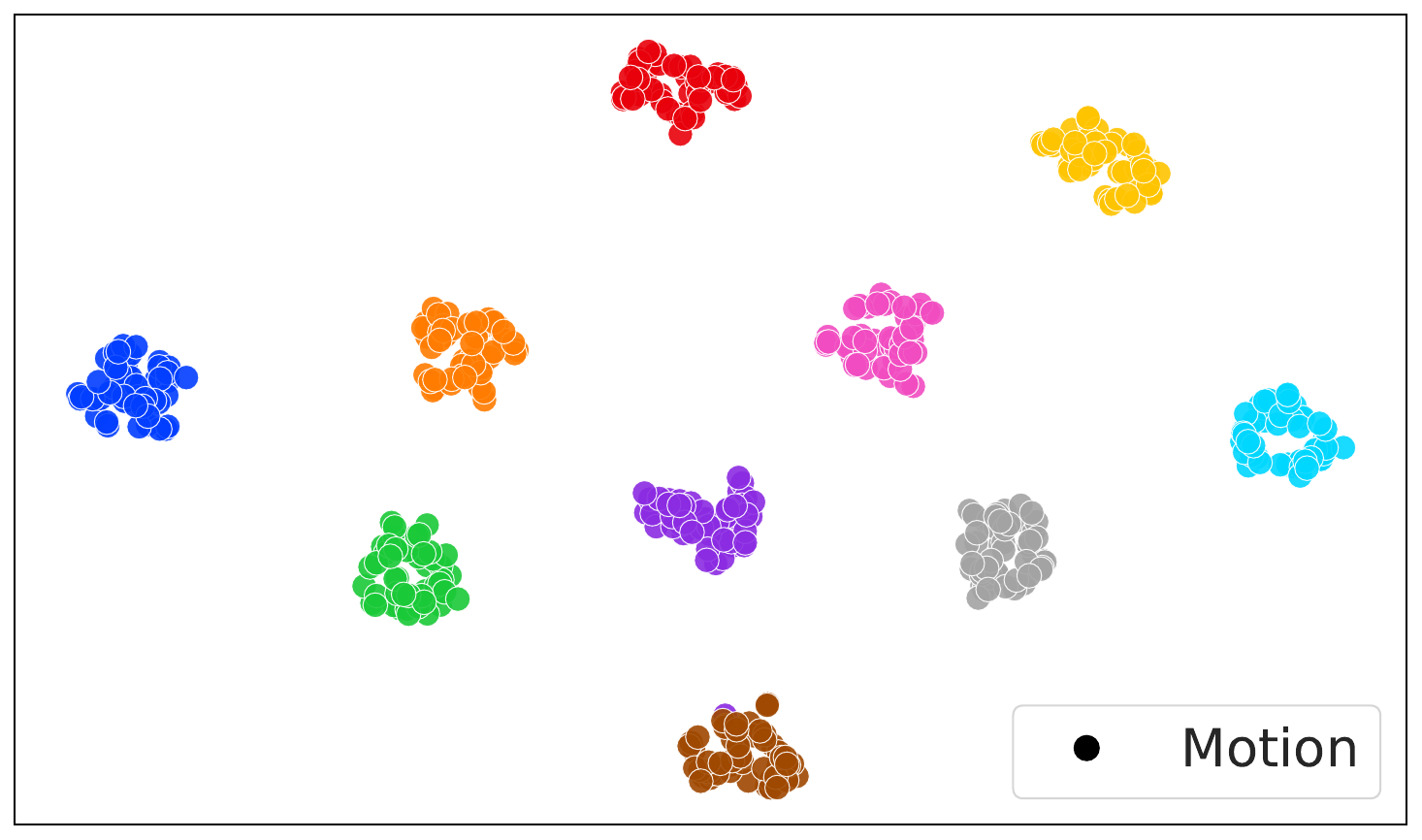}
            \captionof{figure}{\textbf{Feature space of generated motion prompts} across multiple runs (NTU RGB-AB subset).}
            \label{fig:runs}
    \end{minipage}
    \vspace{-4pt}
\end{figure}

\noindent
\textbf{Utility and quality of the generated prompts.}
To asses the impact of accurate textual prompts on disentanglement, we replace non-biometrics descriptions with random clothing details, leading to a $9.2\%$ drop in Rank-1 accuracy on NTU RGB-AB, highlighting the necessity of precise appearance descriptions.
Additionally, we asses prompt consistency by generating descriptions for the same key-frame over five runs on a subset of NTU RGB-AB ($10$ identities and $10$ action classes) and report the average results in Table \ref{tab:runs}. Biometrics descriptions remain highly stable, as indicated by high cosine similarity and low standard deviation, ensuring reliable identity representation. Non-biometrics descriptions also exhibit relative consistency, with minor variations. Motion descriptions exhibit the most variability, as different textual descriptions may be generated for the same action label. Figure \ref{fig:runs} confirms that semantically similar motion prompts still cluster in the same feature space, ensuring consistency in representation.

\noindent
\textbf{Choice of vision encoder and VLM.} Our model supports various vision encoder architectures. To identify the best performer, we evaluated three popular vision encoders: SigLIP-L \cite{zhai2023sigmoid}, ViT-1B from InternVideo2 \cite{wang2024internvideo2}, and ViT-G/14 from EVA-CLIP \cite{sun2023eva} and find ViT-G/14 to be the best performing model (Table \ref{tab:vision_encoder} (\textit{top})). Additionally, we show robustness of our approach across various VLMs, where we observe that changing the VLM does not contribute to significant changes ((Table \ref{tab:vision_encoder} (\textit{bottom}))), thus we select LLaVA for its efficiency and open-source property.

\begin{table}[t!]
    \centering
    \small
    \caption{\textbf{Performance comparison of different vision encoders and VLMs} for NTU RGB-AB and Charades-AB datasets.}
    \begin{tabular}{ll|cc|cc}
    \hline
    \multirow{2}{*}{Model} & \multirow{2}{*}{Size}& \multicolumn{2}{c|}{NTU RGB-AB} & \multicolumn{2}{c}{Charades-AB}\\
    &&R1&mAP&R1&mAP\\
    \hline
    \multicolumn{6}{l}{\textit{  \textcolor{gray}{Vision encoders}}}\\
    SigLIP-L \cite{zhai2023sigmoid}&$0.3$B &$80.2$ & $41.6$ & $48.3$ & $33.7$\\
    ViT-1B \cite{wang2024internvideo2} & $1$B& $\mathbf{83.4}$ & $\underline{42.1}$ & $\underline{49.2}$ & $\underline{34.7}$ \\
    \rowcolor[gray]{.92}
    ViT-G/14 \cite{sun2023eva} &$1.8$B& $\underline{82.2}$ & $\mathbf{43.8}$& $\mathbf{49.9}$ & $\mathbf{34.8}$\\
    \hline
    \multicolumn{5}{l}{\textit{  \textcolor{gray}{Visual Language Models (VLMs)}}}\\
    GPT-4V \cite{openai2024gpt4} &-& $82.3$ & $43.7$ & $49.8$ & $34.9$\\
    InstructBLIP \cite{dai2023instructblip} &$7$B &$82.1$ &$43.8$ & $49.7$ &$34.8$ \\
    \rowcolor[gray]{.92}
    LLaVA 1.5 \cite{liu2023visual}&$7$B&$82.2$ & $43.8$ & $49.9$ & $34.8$\\
    \hline
    \end{tabular}
    \label{tab:vision_encoder}
\end{table}

\begin{figure}
    \centering
    \includegraphics[width=\linewidth]{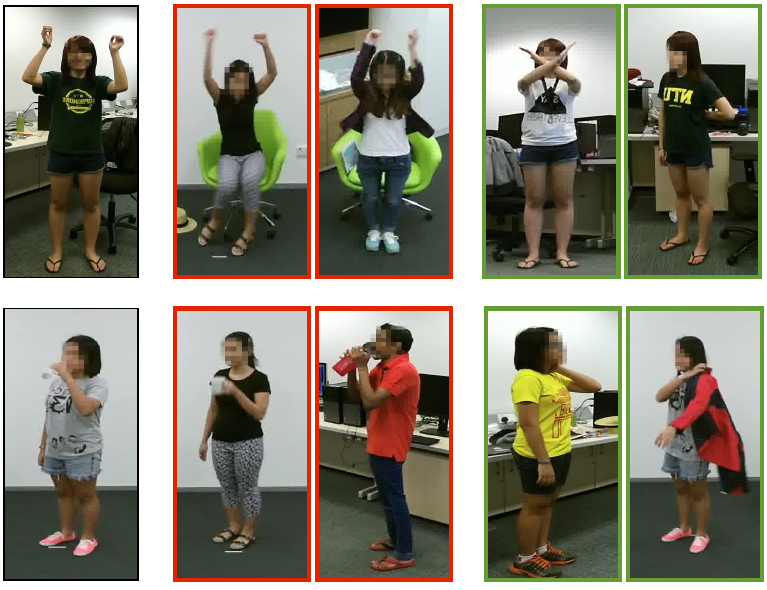}
    \caption{\textbf{Top-2 rank retrieval comparison} of our model and ABNet \cite{azad2024activity}. For a given probe image (\textit{left}), the \textit{middle} two images show incorrect matches retrieved by ABNet due to its over-reliance on activities, while the \textit{right} two images show correct matches retrieved by our model, which effectively disentangles biometrics features from motion cues.}
    \label{fig:qual}
    \vspace{-10pt}
\end{figure}

\noindent
\textbf{Qualitative results.} Figure \ref{fig:qual} compares the top-2 rank retrieval results of our model with ABNet, the only other exisiting activity-biometrics method. As shown, ABNet often misidentifies individuals when they perform the same activity, indicating an over-reliance on motion cues. In contrast, our model, with adaptive motion weighting and effective disentanglement of biometrics, non-biometrics, and motion features, accurately identifies individuals by prioritizing biometrics features over activities. We present more qualitative example in supplementary.

\section{Conclusion}
\label{sec:conclusion}
In this work, we introduce a multimodal \textbf{Disen}tangling \textbf{Q}uerying (\textbf{DisenQ}) Transformer based framework for activity-biometrics, where individuals are identified across diverse activities. Our framework leverages structured language guidance to disentangle biometrics, non-biometrics, and motion cues without additional visual modalities. By performing feature disentanglement, \textbf{DisenQ} ensures that biometrics features remain invariant to non-biometrics information, and motion cues enhance identification without overshadowing stable identity traits. An adaptive weighting mechanism dynamically balances biometrics and motion contributions for reliable identity retrieval. Our approach surpasses existing methods on activity-based benchmarks and generalizes well to traditional video-based identification, demonstrating the importance of effective feature disentanglement in activity-aware person identification.

\section{Acknowledgement}
This research is based upon work primarily supported by the National Science Foundation under Grant No. $2331319$. Any opinions, findings, and conclusions or recommendations contained herein are those of the authors and should not be interpreted as necessarily representing the views, either expressed or implied, of the National Science Foundation.

{
    \small
    \bibliographystyle{ieeenat_fullname}
    \bibliography{main}

\begin{thebibliography}{62}
\providecommand{\natexlab}[1]{#1}
\providecommand{\url}[1]{\texttt{#1}}
\expandafter\ifx\csname urlstyle\endcsname\relax
  \providecommand{\doi}[1]{doi: #1}\else
  \providecommand{\doi}{doi: \begingroup \urlstyle{rm}\Url}\fi

\bibitem[Adjabi et~al.(2020)Adjabi, Ouahabi, Benzaoui, and Taleb-Ahmed]{adjabi2020past}
Insaf Adjabi, Abdeldjalil Ouahabi, Amir Benzaoui, and Abdelmalik Taleb-Ahmed.
\newblock Past, present, and future of face recognition: A review.
\newblock \emph{Electronics}, 9\penalty0 (8):\penalty0 1188, 2020.

\bibitem[Alayrac et~al.(2022)Alayrac, Donahue, Luc, Miech, Barr, Hasson, Lenc, Mensch, Millican, Reynolds, et~al.]{alayrac2022flamingo}
Jean-Baptiste Alayrac, Jeff Donahue, Pauline Luc, Antoine Miech, Iain Barr, Yana Hasson, Karel Lenc, Arthur Mensch, Katherine Millican, Malcolm Reynolds, et~al.
\newblock Flamingo: A visual language model for few-shot learning.
\newblock In \emph{Advances in Neural Information Processing Systems}, pages 23716--23736, 2022.

\bibitem[Azad and Rawat(2024)]{azad2024activity}
Shehreen Azad and Yogesh~Singh Rawat.
\newblock Activity-biometrics: Person identification from daily activities.
\newblock In \emph{CVPR}, pages 287--296, 2024.

\bibitem[Azad et~al.(2025{\natexlab{a}})Azad, Jain, Garg, Vineet, and Rawat]{azad2025understanding}
Shehreen Azad, Yash Jain, Rishit Garg, Vibhav Vineet, and Yogesh Rawat.
\newblock Understanding depth and height perception in large visual-language models.
\newblock In \emph{Proceedings of the Computer Vision and Pattern Recognition Conference}, pages 3611--3620, 2025{\natexlab{a}}.

\bibitem[Azad et~al.(2025{\natexlab{b}})Azad, Vineet, and Rawat]{azad2025hierarq}
Shehreen Azad, Vibhav Vineet, and Yogesh~Singh Rawat.
\newblock Hierarq: Task-aware hierarchical q-former for enhanced video understanding.
\newblock In \emph{Proceedings of the Computer Vision and Pattern Recognition Conference}, pages 8545--8556, 2025{\natexlab{b}}.

\bibitem[Bai et~al.(2022)Bai, Ma, Chang, Huang, and Chen]{bai2022salient}
Shutao Bai, Bingpeng Ma, Hong Chang, Rui Huang, and Xilin Chen.
\newblock Salient-to-broad transition for video person re-identification.
\newblock In \emph{CVPR}, pages 7339--7348, 2022.

\bibitem[Cao et~al.(2022)Cao, Pang, Anwer, Cholakkal, Xie, Shah, and Khan]{cao2022pstr}
Jiale Cao, Yanwei Pang, Rao~Muhammad Anwer, Hisham Cholakkal, Jin Xie, Mubarak Shah, and Fahad~Shahbaz Khan.
\newblock Pstr: End-to-end one-step person search with transformers.
\newblock In \emph{CVPR}, pages 9458--9467, 2022.

\bibitem[Chen et~al.(2020)Chen, Lu, Yang, and Zhou]{chen2020learning}
Guangyi Chen, Jiwen Lu, Ming Yang, and Jie Zhou.
\newblock Learning recurrent 3d attention for video-based person re-identification.
\newblock \emph{IEEE TIP}, 29:\penalty0 6963--6976, 2020.

\bibitem[Chen et~al.(2021)Chen, Jiang, Wang, Zhang, Zheng, Sun, and Zheng]{chen2021learning}
Jiaxing Chen, Xinyang Jiang, Fudong Wang, Jun Zhang, Feng Zheng, Xing Sun, and Wei-Shi Zheng.
\newblock Learning 3d shape feature for texture-insensitive person re-identification.
\newblock In \emph{CVPR}, pages 8146--8155, 2021.

\bibitem[Chen et~al.(2023)Chen, Zhang, Tan, Qu, and Xie]{chen2023unveiling}
Zhong Chen, Zhizhong Zhang, Xin Tan, Yanyun Qu, and Yuan Xie.
\newblock Unveiling the power of clip in unsupervised visible-infrared person re-identification.
\newblock In \emph{Proceedings of the 31st ACM International Conference on Multimedia}, pages 3667--3675, 2023.

\bibitem[Dai et~al.(2023)Dai, Li, Li, Tiong, Zhao, Wang, Li, Fung, and Hoi]{dai2023instructblip}
Wenliang Dai, Junnan Li, Dongxu Li, Anthony Meng~Huat Tiong, Junqi Zhao, Weisheng Wang, Boyang Li, Pascale Fung, and Steven Hoi.
\newblock Instructblip: Towards general-purpose vision-language models with instruction tuning.
\newblock \emph{arXiv preprint arXiv:2305.06500}, 2023.

\bibitem[Davila et~al.(2023)Davila, Du, Lewis, Funk, Van~Pelt, Collins, Corona, Brown, McCloskey, Hoogs, et~al.]{davila2023mevid}
Daniel Davila, Dawei Du, Bryon Lewis, Christopher Funk, Joseph Van~Pelt, Roderic Collins, Kellie Corona, Matt Brown, Scott McCloskey, Anthony Hoogs, et~al.
\newblock Mevid: Multi-view extended videos with identities for video person re-identification.
\newblock In \emph{WACV}, pages 1634--1643, 2023.

\bibitem[Devlin et~al.(2018)Devlin, Chang, Lee, and Toutanova]{devlin2018bert}
Jacob Devlin, Ming-Wei Chang, Kenton Lee, and Kristina Toutanova.
\newblock {BERT}: Pre-training of deep bidirectional transformers for language understanding.
\newblock \emph{arXiv preprint arXiv:1810.04805}, 2018.

\bibitem[Eom and Ham(2019)]{eom2019learning}
Chanho Eom and Bumsub Ham.
\newblock Learning disentangled representation for robust person re-identification.
\newblock \emph{Adv. Neural Inform. Process. Syst.}, 32, 2019.

\bibitem[Eom et~al.(2021)Eom, Lee, Lee, and Ham]{eom2021video}
Chanho Eom, Geon Lee, Junghyup Lee, and Bumsub Ham.
\newblock Video-based person re-identification with spatial and temporal memory networks.
\newblock In \emph{ICCV}, pages 12036--12045, 2021.

\bibitem[Fan et~al.(2020)Fan, Peng, Cao, Liu, Hou, Chi, Huang, Li, and He]{fan2020gaitpart}
Chao Fan, Yunjie Peng, Chunshui Cao, Xu Liu, Saihui Hou, Jiannan Chi, Yongzhen Huang, Qing Li, and Zhiqiang He.
\newblock Gaitpart: Temporal part-based model for gait recognition.
\newblock In \emph{CVPR}, pages 14225--14233, 2020.

\bibitem[Fan et~al.(2023)Fan, Liang, Shen, Hou, Huang, and Yu]{fan2023opengait}
Chao Fan, Junhao Liang, Chuanfu Shen, Saihui Hou, Yongzhen Huang, and Shiqi Yu.
\newblock Opengait: Revisiting gait recognition towards better practicality.
\newblock In \emph{CVPR}, pages 9707--9716, 2023.

\bibitem[Gu et~al.(2020)Gu, Chang, Ma, Zhang, and Chen]{gu2020appearance}
Xinqian Gu, Hong Chang, Bingpeng Ma, Hongkai Zhang, and Xilin Chen.
\newblock Appearance-preserving 3d convolution for video-based person re-identification.
\newblock In \emph{ECCV}, pages 228--243. Springer, 2020.

\bibitem[Gu et~al.(2022)Gu, Chang, Ma, Bai, Shan, and Chen]{gu2022clothes}
Xinqian Gu, Hong Chang, Bingpeng Ma, Shutao Bai, Shiguang Shan, and Xilin Chen.
\newblock Clothes-changing person re-identification with rgb modality only.
\newblock In \emph{CVPR}, pages 1060--1069, 2022.

\bibitem[Guo et~al.(2023)Guo, Liu, Wu, Wang, and Wang]{guo2023SCNet}
Peini Guo, Hong Liu, Jianbing Wu, Guoquan Wang, and Tao Wang.
\newblock Semantic-aware consistency network for cloth-changing person re-identification.
\newblock In \emph{ACM MM}, 2023.

\bibitem[He et~al.(2021)He, Jin, Shen, Huang, Chen, and Hua]{he2021dense}
Tianyu He, Xin Jin, Xu Shen, Jianqiang Huang, Zhibo Chen, and Xian-Sheng Hua.
\newblock Dense interaction learning for video-based person re-identification.
\newblock In \emph{Proceedings of the IEEE/CVF International Conference on Computer Vision}, pages 1490--1501, 2021.

\bibitem[He et~al.(2024)He, Deng, Tang, Chen, Xie, Wang, Bai, Zhu, Zhao, Ouyang, et~al.]{he2024instruct}
Weizhen He, Yiheng Deng, Shixiang Tang, Qihao Chen, Qingsong Xie, Yizhou Wang, Lei Bai, Feng Zhu, Rui Zhao, Wanli Ouyang, et~al.
\newblock Instruct-reid: A multi-purpose person re-identification task with instructions.
\newblock In \emph{CVPR}, pages 17521--17531, 2024.

\bibitem[Hong et~al.(2021)Hong, Wu, Wu, Han, and Zheng]{hong2021fine}
Peixian Hong, Tao Wu, Ancong Wu, Xintong Han, and Wei-Shi Zheng.
\newblock Fine-grained shape-appearance mutual learning for cloth-changing person re-identification.
\newblock In \emph{CVPR}, pages 10513--10522, 2021.

\bibitem[Hou et~al.(2020)Hou, Chang, Ma, Shan, and Chen]{hou2020temporal}
Ruibing Hou, Hong Chang, Bingpeng Ma, Shiguang Shan, and Xilin Chen.
\newblock Temporal complementary learning for video person re-identification.
\newblock In \emph{ECCV}, pages 388--405. Springer, 2020.

\bibitem[Hou et~al.(2021)Hou, Chang, Ma, Huang, and Shan]{hou2021bicnet}
Ruibing Hou, Hong Chang, Bingpeng Ma, Rui Huang, and Shiguang Shan.
\newblock Bicnet-tks: Learning efficient spatial-temporal representation for video person re-identification.
\newblock In \emph{CVPR}, pages 2014--2023, 2021.

\bibitem[Hu et~al.(2024)Hu, Yang, and Ye]{2024TVI-LFM}
Zhangyi Hu, Bin Yang, and Mang Ye.
\newblock Empowering visible-infrared person re-identification with large foundation models.
\newblock In \emph{NeurIPS}, 2024.

\bibitem[Huang et~al.(2019)Huang, Wu, Xu, and Zhong]{huang2019celebrities}
Yan Huang, Qiang Wu, Jingsong Xu, and Yi Zhong.
\newblock Celebrities-reid: A benchmark for clothes variation in long-term person re-identification.
\newblock In \emph{IJCNN}, pages 1--8. IEEE, 2019.

\bibitem[Jiang et~al.(2020)Jiang, Gong, Guo, Yang, Huang, Zheng, Zheng, and Sun]{jiang2020rethinking}
Xinyang Jiang, Yifei Gong, Xiaowei Guo, Qize Yang, Feiyue Huang, Wei-Shi Zheng, Feng Zheng, and Xing Sun.
\newblock Rethinking temporal fusion for video-based person re-identification on semantic and time aspect.
\newblock In \emph{AAAI}, pages 11133--11140, 2020.

\bibitem[Jin et~al.(2022)Jin, He, Zheng, Yin, Shen, Huang, Feng, Huang, Chen, and Hua]{jin2022cloth}
Xin Jin, Tianyu He, Kecheng Zheng, Zhiheng Yin, Xu Shen, Zhen Huang, Ruoyu Feng, Jianqiang Huang, Zhibo Chen, and Xian-Sheng Hua.
\newblock Cloth-changing person re-identification from a single image with gait prediction and regularization.
\newblock In \emph{CVPR}, pages 14278--14287, 2022.

\bibitem[Li et~al.(2024)Li, Zhang, Guo, Zhang, Li, Zhang, Zhang, Li, Liu, and Li]{li2024llava}
Bo Li, Yuanhan Zhang, Dong Guo, Renrui Zhang, Feng Li, Hao Zhang, Kaichen Zhang, Yanwei Li, Ziwei Liu, and Chunyuan Li.
\newblock Llava-onevision: Easy visual task transfer.
\newblock \emph{arXiv preprint arXiv:2408.03326}, 2024.

\bibitem[Li et~al.(2023{\natexlab{a}})Li, Li, Savarese, and Hoi]{li2023blip}
Junnan Li, Dongxu Li, Silvio Savarese, and Steven Hoi.
\newblock Blip-2: Bootstrapping language-image pre-training with frozen image encoders and large language models.
\newblock In \emph{ICML}, pages 19730--19742. PMLR, 2023{\natexlab{a}}.

\bibitem[Li et~al.(2023{\natexlab{b}})Li, Sun, and Li]{li2023clip}
Siyuan Li, Li Sun, and Qingli Li.
\newblock Clip-reid: exploiting vision-language model for image re-identification without concrete text labels.
\newblock In \emph{AAAI}, pages 1405--1413, 2023{\natexlab{b}}.

\bibitem[Liang et~al.(2022)Liang, Fan, Hou, Shen, Huang, and Yu]{liang2022gaitedge}
Junhao Liang, Chao Fan, Saihui Hou, Chuanfu Shen, Yongzhen Huang, and Shiqi Yu.
\newblock Gaitedge: Beyond plain end-to-end gait recognition for better practicality.
\newblock In \emph{ECCV}, pages 375--390. Springer, 2022.

\bibitem[Liang and Rawat(2025)]{liang2025differ}
Xin Liang and Yogesh~S Rawat.
\newblock Differ: Disentangling identity features via semantic cues for clothes-changing person re-id.
\newblock In \emph{Proceedings of the Computer Vision and Pattern Recognition Conference}, pages 13980--13989, 2025.

\bibitem[Lin et~al.(2021)Lin, Zhang, and Yu]{lin2021gait}
Beibei Lin, Shunli Zhang, and Xin Yu.
\newblock Gait recognition via effective global-local feature representation and local temporal aggregation.
\newblock In \emph{ICCV}, pages 14648--14656, 2021.

\bibitem[Liu et~al.(2023{\natexlab{a}})Liu, Li, Wu, and Lee]{liu2023visual}
Haotian Liu, Chunyuan Li, Qingyang Wu, and Yong~Jae Lee.
\newblock Visual instruction tuning.
\newblock \emph{NeurIPS}, 36:\penalty0 34892--34916, 2023{\natexlab{a}}.

\bibitem[Liu et~al.(2023{\natexlab{b}})Liu, Yu, Zhang, and Lu]{liu2023deeply}
Xuehu Liu, Chenyang Yu, Pingping Zhang, and Huchuan Lu.
\newblock Deeply coupled convolution--transformer with spatial--temporal complementary learning for video-based person re-identification.
\newblock \emph{IEEE Transactions on Neural Networks and Learning Systems}, 2023{\natexlab{b}}.

\bibitem[Meng et~al.(2021)Meng, Zhao, Huang, and Zhou]{meng2021magface}
Qiang Meng, Shichao Zhao, Zhida Huang, and Feng Zhou.
\newblock Magface: A universal representation for face recognition and quality assessment.
\newblock In \emph{CVPR}, pages 14225--14234, 2021.

\bibitem[OpenAI(2024)]{openai2024gpt4}
OpenAI.
\newblock Gpt-4 technical report.
\newblock \emph{arXiv preprint arXiv:2303.08774}, 2024.

\bibitem[Pathak(2020)]{pathak2020fine}
Priyank Pathak.
\newblock Fine-grained re-identification.
\newblock \emph{arXiv preprint arXiv:2011.13475}, 2020.

\bibitem[Pathak and Rawat(2025{\natexlab{a}})]{Pathak_2025_ICCV}
Priyank Pathak and Yogesh~S Rawat.
\newblock Colors see colors ignore: Clothes changing reid with color disentanglement.
\newblock In \emph{Proceedings of the IEEE/CVF International Conference on Computer Vision (ICCV)}, 2025{\natexlab{a}}.

\bibitem[Pathak and Rawat(2025{\natexlab{b}})]{pathak2025coarse}
Priyank Pathak and Yogesh~S Rawat.
\newblock Coarse attribute prediction with task agnostic distillation for real world clothes changing reid.
\newblock \emph{arXiv preprint arXiv:2505.12580}, 2025{\natexlab{b}}.

\bibitem[Pathak et~al.(2020)Pathak, Eshratifar, and Gormish]{pathak2020video}
Priyank Pathak, Amir~Erfan Eshratifar, and Michael Gormish.
\newblock Video person re-id: Fantastic techniques and where to find them (student abstract).
\newblock In \emph{AAAI}, pages 13893--13894, 2020.

\bibitem[Porrello et~al.(2020)Porrello, Bergamini, and Calderara]{porrello2020robust}
Angelo Porrello, Luca Bergamini, and Simone Calderara.
\newblock Robust re-identification by multiple views knowledge distillation.
\newblock In \emph{ECCV}, pages 93--110. Springer, 2020.

\bibitem[Qian et~al.(2020)Qian, Wang, Zhang, Zhu, Fu, Xiang, Jiang, and Xue]{qian2020long}
Xuelin Qian, Wenxuan Wang, Li Zhang, Fangrui Zhu, Yanwei Fu, Tao Xiang, Yu-Gang Jiang, and Xiangyang Xue.
\newblock Long-term cloth-changing person re-identification.
\newblock In \emph{ACCV}, 2020.

\bibitem[Radford et~al.(2021)Radford, Kim, Hallacy, Ramesh, Goh, Agarwal, Sastry, Askell, Mishkin, Clark, Krueger, and Sutskever]{radford2021learningtransferablevisualmodels}
Alec Radford, Jong~Wook Kim, Chris Hallacy, Aditya Ramesh, Gabriel Goh, Sandhini Agarwal, Girish Sastry, Amanda Askell, Pamela Mishkin, Jack Clark, Gretchen Krueger, and Ilya Sutskever.
\newblock Learning transferable visual models from natural language supervision, 2021.

\bibitem[Schiappa et~al.(2024{\natexlab{a}})Schiappa, Abdullah, Azad, Claypoole, Cogswell, Divakaran, and Rawat]{schiappa2024probing}
Madeline Schiappa, Raiyaan Abdullah, Shehreen Azad, Jared Claypoole, Michael Cogswell, Ajay Divakaran, and Yogesh Rawat.
\newblock Probing conceptual understanding of large visual-language models.
\newblock In \emph{Proceedings of the IEEE/CVF Conference on Computer Vision and Pattern Recognition}, pages 1797--1807, 2024{\natexlab{a}}.

\bibitem[Schiappa et~al.(2024{\natexlab{b}})Schiappa, Azad, Vs, Ge, Miksik, Rawat, and Vineet]{schiappa2024robustness}
Madeline~Chantry Schiappa, Shehreen Azad, Sachidanand Vs, Yunhao Ge, Ondrej Miksik, Yogesh~S Rawat, and Vibhav Vineet.
\newblock Robustness analysis on foundational segmentation models.
\newblock In \emph{Proceedings of the IEEE/CVF Conference on Computer Vision and Pattern Recognition}, pages 1786--1796, 2024{\natexlab{b}}.

\bibitem[Sun et~al.(2023)Sun, Fang, Wu, Wang, and Cao]{sun2023eva}
Quan Sun, Yuxin Fang, Ledell Wu, Xinlong Wang, and Yue Cao.
\newblock Eva-clip: Improved training techniques for clip at scale.
\newblock \emph{arXiv preprint arXiv:2303.15389}, 2023.

\bibitem[Wang et~al.(2024{\natexlab{a}})Wang, Li, and Xue]{wang2024large}
Qizao Wang, Bin Li, and Xiangyang Xue.
\newblock When large vision-language models meet person re-identification.
\newblock \emph{arXiv preprint arXiv:2411.18111}, 2024{\natexlab{a}}.

\bibitem[Wang et~al.(2021)Wang, Zhang, Gao, Geng, Lu, and Wang]{wang2021pyramid}
Yingquan Wang, Pingping Zhang, Shang Gao, Xia Geng, Hu Lu, and Dong Wang.
\newblock Pyramid spatial-temporal aggregation for video-based person re-identification.
\newblock In \emph{ICCV}, pages 12026--12035, 2021.

\bibitem[Wang et~al.(2024{\natexlab{b}})Wang, Li, Li, Yu, He, Chen, Pei, Zheng, Wang, Shi, et~al.]{wang2024internvideo2}
Yi Wang, Kunchang Li, Xinhao Li, Jiashuo Yu, Yinan He, Guo Chen, Baoqi Pei, Rongkun Zheng, Zun Wang, Yansong Shi, et~al.
\newblock Internvideo2: Scaling foundation models for multimodal video understanding.
\newblock In \emph{ECCV}, pages 396--416. Springer, 2024{\natexlab{b}}.

\bibitem[Yan et~al.(2023)Yan, Dong, Zhang, and Tang]{yan2023clip}
Shuanglin Yan, Neng Dong, Liyan Zhang, and Jinhui Tang.
\newblock Clip-driven fine-grained text-image person re-identification.
\newblock \emph{IEEE Transactions on Image Processing}, 32:\penalty0 6032--6046, 2023.

\bibitem[Yang et~al.(2019)Yang, Wu, and Zheng]{yang2019person}
Qize Yang, Ancong Wu, and Wei-Shi Zheng.
\newblock Person re-identification by contour sketch under moderate clothing change.
\newblock \emph{IEEE TPAMI}, 43\penalty0 (6):\penalty0 2029--2046, 2019.

\bibitem[Yang et~al.(2023)Yang, Lin, Zhong, Wu, and Wang]{yang2023good}
Zhengwei Yang, Meng Lin, Xian Zhong, Yu Wu, and Zheng Wang.
\newblock Good is bad: Causality inspired cloth-debiasing for cloth-changing person re-identification.
\newblock In \emph{CVPR}, pages 1472--1481, 2023.

\bibitem[Yang et~al.(2024)Yang, Wu, Wu, Lin, Gu, and Wang]{yang2024pedestrian}
Zexian Yang, Dayan Wu, Chenming Wu, Zheng Lin, Jingzi Gu, and Weiping Wang.
\newblock A pedestrian is worth one prompt: Towards language guidance person re-identification.
\newblock In \emph{Proceedings of the IEEE/CVF Conference on Computer Vision and Pattern Recognition}, pages 17343--17353, 2024.

\bibitem[Ye et~al.(2021)Ye, Shen, Lin, Xiang, Shao, and Hoi]{ye2021deep}
Mang Ye, Jianbing Shen, Gaojie Lin, Tao Xiang, Ling Shao, and Steven~CH Hoi.
\newblock Deep learning for person re-identification: A survey and outlook.
\newblock \emph{IEEE TPAMI}, 44\penalty0 (6):\penalty0 2872--2893, 2021.

\bibitem[Yu et~al.(2024)Yu, Liu, Wang, Zhang, and Lu]{yu2024tf}
Chenyang Yu, Xuehu Liu, Yingquan Wang, Pingping Zhang, and Huchuan Lu.
\newblock Tf-clip: Learning text-free clip for video-based person re-identification.
\newblock In \emph{AAAI}, pages 6764--6772, 2024.

\bibitem[Zang et~al.(2022)Zang, Li, and Gao]{zang2022multidirection}
Xianghao Zang, Ge Li, and Wei Gao.
\newblock Multidirection and multiscale pyramid in transformer for video-based pedestrian retrieval.
\newblock \emph{IEEE Transactions on Industrial Informatics}, 18\penalty0 (12):\penalty0 8776--8785, 2022.

\bibitem[Zhai et~al.(2023)Zhai, Mustafa, Kolesnikov, and Beyer]{zhai2023sigmoid}
Xiaohua Zhai, Basil Mustafa, Alexander Kolesnikov, and Lucas Beyer.
\newblock Sigmoid loss for language image pre-training.
\newblock In \emph{Proceedings of the IEEE/CVF international conference on computer vision}, pages 11975--11986, 2023.

\bibitem[Zhu et~al.(2023)Zhu, Chen, Shen, Li, and Elhoseiny]{zhu2023minigpt4}
Deyao Zhu, Jun Chen, Xiaoqian Shen, Xiang Li, and Mohamed Elhoseiny.
\newblock Minigpt-4: Enhancing vision-language understanding with advanced large language models.
\newblock \emph{arXiv preprint arXiv:2304.10592}, 2023.

\bibitem[Zhu et~al.(2024)Zhu, Zheng, Zheng, and Nevatia]{zhu2024sharc}
Haidong Zhu, Wanrong Zheng, Zhaoheng Zheng, and Ram Nevatia.
\newblock Sharc: Shape and appearance recognition for person identification in-the-wild.
\newblock In \emph{WACV}, pages 6290--6300, 2024.

\end{thebibliography}
}

\clearpage
\setcounter{page}{1}
\renewcommand{\thesection}{\Alph{section}}
\setcounter{section}{0}
\maketitlesupplementary
\begin{figure*}[t!]
    \centering
    \includegraphics[width=\linewidth]{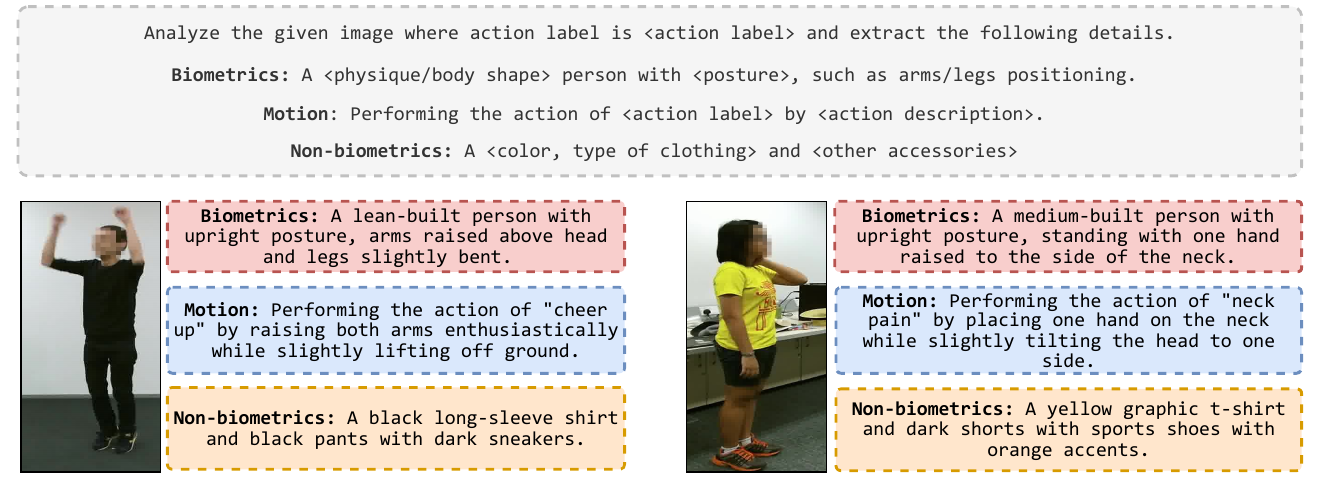}
    \caption{\textbf{Examples of generated structured textual descriptions.}}
    \label{fig:prompt}
\end{figure*}

In this supplementary material, we provide details of prompt generation in Section \ref{sec:prompt_su}, along with structured examples of generated descriptions. Then we present additional quantitative results and analysis in Section \ref{sec:res_sup} and Section \ref{sec:ana_sup}, followed by qualitative analysis in Section \ref{sec:res_qual}. Then we present the dataset statistics in Section \ref{sec:data}. Finally, we address some limitations and outline directions for future research in Section \ref{sec:future}.

\section{Details of Prompt Generation}
\label{sec:prompt_su}
We use the following structured prompt template designed to extract biometrics, non-biometrics, and motion-related details from the key-frame of a given RGB video.

\begin{quote}
    \texttt{Analyze the given image where action label is <action label> and extract the following details: 
    Biometrics: A <physique/body shape> person with <posture>, such as arms/legs positioning.
    Motion: Performing the action of <action label> by <action description>.
    Non-biometrics: A <color, type of clothing> and <other accessories>.}
\end{quote}

This prompt template is fed into the frozen VLM along with the key-frame, allowing the model to generate structured textual descriptions for each feature category. The output is then parsed into three distinct textual embeddings corresponding to biometrics, motion, and non-biometrics, ensuring explicit separation of identity-related and appearance-based cues. By incorporating structured textual supervision, this approach enhances feature disentanglement, enabling the model to learn identity-relevant representations while mitigating appearance bias. In Figure \ref{fig:prompt}, we present examples of structured textual descriptions generated using a Vision-Language Model (VLM) from a given key-frame and its associated action label.
\label{sec:prompt}

\section{Additional Results}
\label{sec:res_sup}
In Table \ref{tab:r5} we present performance comparison of our method with existing works and report the rank 5 accuracy. We present the result of our model on the excluding same view evaluation protocol in Table \ref{tab:comprehensive_eval}. From both of these tables, we observe that our model constantly outperforms all the existing models across all datasets. 

\begin{table*}[t!]
    \centering

    \caption{\textbf{Performance comparison of person reidentification} on NTU RGB-AB, PKU MMD-AB, and Charades-AB datasets. Here we report rank 5 accuracies. $\dag$ represents results produced in our environment. \textbf{Bold} represents best results.}

    \begin{tabular}{ll|cc|cc|cc}
        \hline
        \multirow{2}{*}{Model} & \multirow{2}{*}{Venue} & \multicolumn{2}{c|}{NTU RGB-AB} & \multicolumn{2}{c|}{PKU MMD-AB} & \multicolumn{2}{c}{Charades-AB} \\
        & & Same & Cross & Same & Cross & Same & Cross \\
        \hline
        \multicolumn{8}{l}{\textit{ \textcolor{gray}{Models with only visual modality}}}\\
        TSF \cite{jiang2020rethinking} & AAAI 20 & $72.9$ & $70.3$ & $78.5$ & $73.5$ & $38.2$ & $32.1$ \\
        VKD \cite{porrello2020robust} & ECCV 20 & $68.9$ & $69.2$ & $80.0$ & $74.3$ & $38.9$ & $34.4$ \\
        BiCnet-TKS \cite{hou2021bicnet} & CVPR 21 & $75.7$ & $70.7$ & $83.0$ & $78.7$ & $41.9$ & $40.6$ \\
        PSTA \cite{wang2021pyramid} & ICCV 21 & $69.7$ & $67.7$ & $79.1$ & $74.0$ & $45.0$ & $40.5$ \\
        STMN \cite{eom2021video} & ICCV 21 & $74.8$ & $71.9$ & $79.6$ & $73.3$ & $41.3$ & $35.3$ \\
        SINet \cite{bai2022salient} & CVPR 22 & $71.1$ & $69.1$ & $82.2$ & $78.0$ & $42.3$ & $38.7$ \\
        CAL \cite{gu2022clothes} & CVPR 22 & $78.6$ & $76.5$ & $86.0$ & $81.2$ & $48.2$ & $45.3$ \\
        Video-CAL \cite{gu2022clothes} & CVPR 22 & $81.3$ & $79.5$ & $83.1$ & $82.5$ & $50.1$ & $48.5$ \\
        PSTR \cite{cao2022pstr} & CVPR 22& $71.2$  &$69.3$& $85.2$ &$80.0$& $40.2$ & $37.2$ \\
        AIM \cite{yang2023good} & CVPR 23& $73.4$ & $71.8$& ${83.5}$  &$80.4$& $42.1$ & $37.6$ \\
        SCNet \cite{guo2023SCNet} & ACM MM 23& $71.9$  &$70.3$&$81.4$ &$74.9$& $34.5$ &$30.2$ \\
       
        ABNet \cite{azad2024activity} & CVPR 24 & $85.3$ & $81.4$ & $91.4$ & $89.3$ & $51.0$ & $52.0$ \\
        \hline\hline
        \multicolumn{8}{l}{\textit{ \textcolor{gray}{Models with visual +language modality}}}\\
        CLIP ReID $\dag$ \cite{li2023clip} & AAAI 23 & $79.2$ & $77.3$ & $85.0$ & $83.2$ & $46.8$ & $44.6$ \\
        CCLNet $\dag$ \cite{chen2023unveiling} & ACM MM 23 & $78.2$ & $77.1$ & $86.7$ & $82.5$ & $45.9$ & $41.7$ \\
        TF-CLIP $\dag$ \cite{yu2024tf} & AAAI 24 & $79.6$ & $77.0$ & $85.9$ & $84.1$ & $43.7$ & $42.1$ \\
        TVI-LFM $\dag$ \cite{2024TVI-LFM} & NeurIPS 24 & $78.9$ & $77.5$ & $87.1$ & $83.5$ & $49.5$ & $46.3$ \\
        Instruct-ReID $\dag$ \cite{he2024instruct} & CVPR 24 & $81.1$ & $79.6$ & $87.3$ & $83.5$ & $47.9$ & $43.1$ \\
        \hline
        EVA-CLIP \cite{sun2023eva} & & $75.4$ & $72.8$ & $77.2$ & $72.1$ & $41.3$ & $33.8$ \\
        \rowcolor[gray]{.92} Ours & & $\mathbf{88.5}$ & $\mathbf{86.4}$ & $\mathbf{94.7}$ & $\mathbf{90.5}$ & $\mathbf{56.8}$ & $\mathbf{54.1}$ \\
        \hline
    \end{tabular}

    \label{tab:r5}
\end{table*}

\begin{table}[t!]

    \centering
    \caption{Performance comparison of our model on NTU RGB-AB, PKU MMD-AB, and Charades-AB datasets for excluding same view evaluation protocols.}

    \begin{tabular}{l|p{1cm}|l|c|c}
        \hline
         & Eval. & Model & Rank 1 & mAP\\
        \hline
        \multirow{4}{*}{NTU} & \multirow{2}{.5cm}{Same activity} & ABNet \cite{azad2024activity} & $77.8$ & $38.8$\\
        && DisenQ & $\mathbf{80.7}$ & $\mathbf{40.9}$\\
        \cline{2-5}
        & \multirow{2}{.5cm}{Cross activity} & ABNet \cite{azad2024activity} & $76.4$ & $36.1$\\
        && DisenQ & $\mathbf{79.3}$ & $\mathbf{37.6}$\\
        \hline
        \multirow{4}{*}{PKU} & \multirow{2}{.5cm}{Same activity} & ABNet \cite{azad2024activity} & $81.4$ & $51.7$\\
        && DisenQ & $\mathbf{84.2}$ & $\mathbf{55.1}$\\
        \cline{2-5}
        & \multirow{2}{.5cm}{Cross activity} & ABNet \cite{azad2024activity} & $79.4$ & $46.3$\\
        && DisenQ & $\mathbf{82.4}$ & $\mathbf{50.5}$\\
        \hline
    \end{tabular}

    \label{tab:comprehensive_eval}

    \end{table}

\section{Additional Analysis}
\label{sec:ana_sup}
\subsection{VLM quality is not a performance scalability bottleneck}
\begin{figure}[t!]
    \centering
    \includegraphics[width=.8\linewidth]{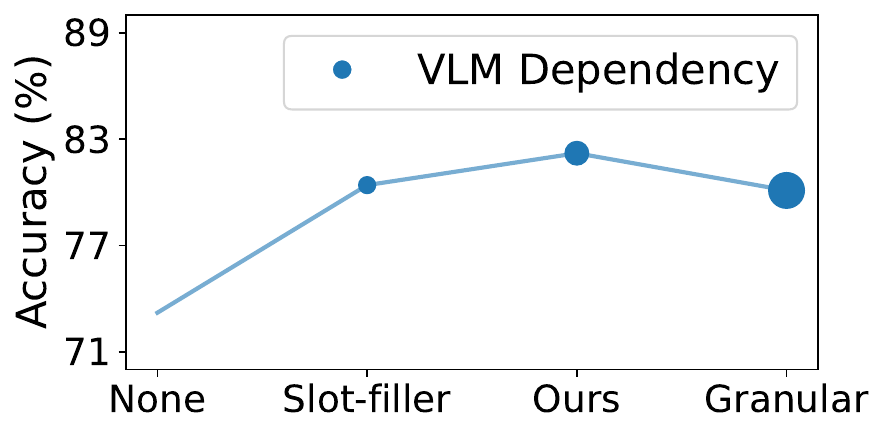}
    \caption{Prompt structure vs. accuracy and VLM reliance (NTU RGB-AB).}
    \label{fig:vlm}
\end{figure}

We evaluate VLM scalability using two prompt variants: simplified slot-filler prompts (e.g., “\texttt{a [body shape] person}”), and fine-grained 4-way disentanglement (biometrics, motion, (upper/lower)-body clothing). As shown in  Figure~\ref{fig:vlm},  even simple prompts significantly improve performance over no prompts, highlighting the value of semantic structure over linguistic richness. 
In contrast, granular prompts reduce performance and add complexity. Since all fine-grained features fall within the core axes of biometrics, non-biometrics, and motion, our 3-way setup remains more robust and scalable. Moreover, by restricting VLM use to \textit{training only} and using structured prompts, we reduce noise and ensure that performance does not heavily depend on VLM strength.

\subsection{Cross-domain utility of disentangled features.}
We evaluated cross-domain generalization from NTU RGB-AB to PKU MMD-AB (Table~\ref{tab:tab2}), and found that while biometrics ($F_b$) transfer well, motion ($F_m$) showed slightly lower performance due to action variability. Their combination yields the best performance, confirming their complementary strengths (Table \ref{tab:abla_main}) even across domains.  
\begin{table}
    \centering
    \caption{\textbf{Utility of disentangled features} across domain.}
    \begin{tabular}{l|c}
    \hline
         Feature & Rank 1  \\
         \hline
         Baseline & $61.7$\\
         $F_b$ & $74.3$\\
         $F_m$& $68.1$\\
         \rowcolor[gray]{.92}
         $F_b$, $F_m$ & $\mathbf{76.8}$\\
         \hline
    \end{tabular}

    \label{tab:tab2}
\end{table}

\subsection{Further analyzing feature disentanglement} We conducted three analyses to validate that the model truly separates features in a visually grounded manner—rather than merely aligning to prompt format.

\begin{figure} [t!]
    \centering
    \includegraphics[width=.8\linewidth]{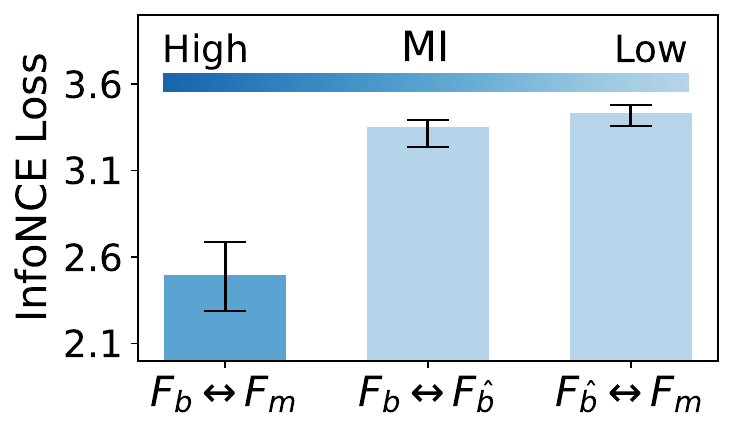}
    \caption{Mutual info (MI) analysis.}
    \label{fig:infonce_mi}
\end{figure}

\noindent\textbf{Mutual Information Analysis:} To verify disentanglement, we compute InfoNCE-based mutual information between each feature pair using empirically estimated upper and lower bounds derived from matched (same actor/action) and mismatched (different actor/action) pairs of NTU RGB-AB (Figure~\ref{fig:infonce_mi}). The relatively lower InfoNCE for $F_b \leftrightarrow F_m$ falls within a wider range, indicating some mutual information, expected due to identity-linked motion cues such as gait, swing style etc. This highlights $F_m$ as a complementary cue to $F_b$ for identity matching. In contrast, $F_b \leftrightarrow F_{\hat{b}}$ and $F_m \leftrightarrow F_{\hat{b}}$ show consistently higher losses within tighter bounds, confirming minimal shared information and effective disentanglement.

\noindent\textbf{Cross Feature Leakage Test:}
To further verify disentanglement, we trained classifiers to predict action from $F_b$ (biometrics) and $F_{\hat{b}}$ (non-biometrics) on NTU RGB-AB. $F_b$ achieved $14.4$\% accuracy, reflecting some posture-related cues embedded in body shape. These cues are expected, as they are stable biometrics components, but do not represent dynamic motion. On the contrary, $F_{\hat{b}}$ scored near random ($4.8$\%), confirming no motion leakage into appearance features. These results supports minimal unintended information transfer across branches.
\noindent
\textbf{Causal Intervention:}
Additionally, to test if disentanglement stems from prompt structure, we swapped prompt semantics across branches without changing losses. Despite this deliberate mismatch (e.g., motion prompts guiding biometrics features), identity and action performance dropped only marginally ($1–2$\%) on NTU RGB-AB, suggesting that feature separation is guided by visual supervision rather than prompt formatting.

\subsection{Risk of VLMs' inherent bias propagation} To address potential VLM bias linking appearance with identity,  we use the VLM only to generate controlled attributes' descriptions within predefined, structured prompt templates, not free-form text. We further mitigate residual VLM correlations by enforcing orthogonality (Equation \ref{eq:Orth}) and excluding non-biometrics features from identity matching (Equation \ref{eq:total_loss}). Together, these steps minimize any implicit bias and maintain clean disentanglement between appearance and identity features. While these design choices aim to mitigate potential sources of bias, we acknowledge that some demographic bias may still persist due to upstream VLM pretraining, which is beyond the scope of this work.

\subsection{Non-biometrics branch encodes appearance information to some extent}
While the non-biometrics branch lacks explicit supervision, it is guided by appearance-focused prompts and regularized via orthogonality to remain distinct from biometrics. Color histogram analysis of Figure \ref{fig:tsne} indicates that neighboring pairs in the non-biometrics space tend to have more similar appearance attributes ($0.81$ vs. $0.54$ for random pairs), suggesting that the learned features in this branch reflect clothing-related information to some extent.

\section{Qualitative Results}
\label{sec:res_qual}
\begin{figure}
    \centering
    \includegraphics[width=\linewidth]{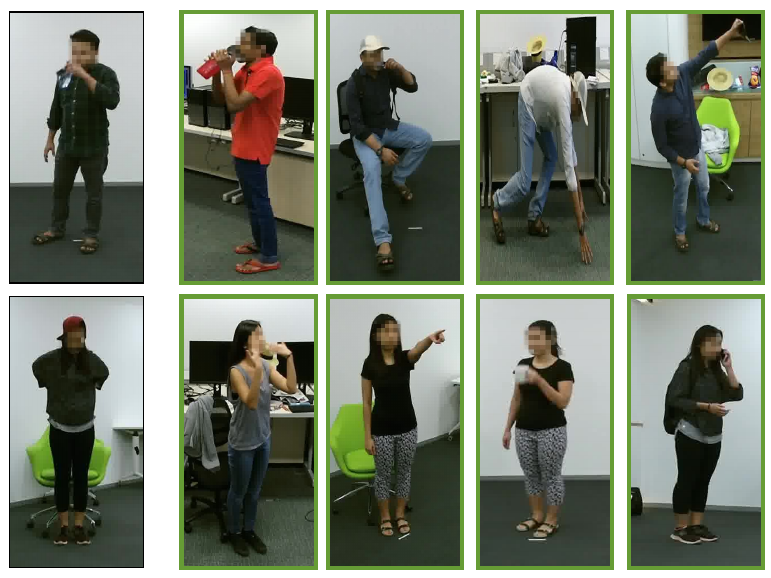}
    \caption{\textbf{Qualitative results}. Here we present the top 4 rank retrieval results for a given probe (left) of our model on same-activity (\textit{top)} and cross-activity (\textit{bottom)} on NTU RGB-AB dataset.}
    \label{fig:ntu}
\end{figure}

Figure \ref{fig:ntu} illustrates the top 4 rank retrieval results for a given probe for NTU RGB-AB dataset in both same and cross-activity evaluation setting. This demonstrates the robustness of our model across diverse activities and significant appearance variations. Unlike traditional approaches that struggle with identity retention under clothing changes or motion variations, our method effectively disentangles biometrics, non-biometrics, and motion cues, ensuring accurate identification even when activities differ between the probe and gallery. The strong retrieval performance highlights the effectiveness of our approach in learning identity-consistent representations that generalize across diverse set of real-world activities.

\section{Dataset statistics}
\label{sec:data}
We evaluate performance under two evaluation protocols: same-activity and cross-activity. In the same-activity setting, all activities are present across both sets, ensuring that each individual is observed performing the same set of actions. In contrast, the cross-activity protocol introduces a more challenging scenario where individuals appear in different activities across the two sets, meaning that activities seen in one set are entirely absent in the other.  
For datasets with multiple viewpoints, such as NTU RGB-AB and PKU MMD-AB, we further assess two variations: including same view, where all viewpoints are available in both probe and gallery, and excluding same view, where probe viewpoint is excluded from gallery, increasing the difficulty of matching individuals across different perspectives. This allows us to analyze the model’s robustness to viewpoint variations. However, for datasets like Charades-AB, which do not contain explicit viewpoints data, only the activity-based protocols are considered. Since, MEVID only contains one activity (e.g. walking), the evaluation of this dataset also falls under the same-activity setting.  As MEVID primarily features walking sequences, we assign all tracklets a ``walking" label to enable coarse motion supervision, while being consistent with standard re-ID protocols that leverage gait. Since occasional secondary actions are concurrent with walking, it allows us to still use motion supervision without explicit activity labels. A detailed dataset statistics is presented in Table \ref{tab:data_description}.

\begin{table}[t!]
    \centering
    \small
    \caption{\textbf{{Dataset statistics}}}
    \begin{tabular}{p{1.7cm}l|ccc}
         \hline
         Dataset& Split&\#actors&	\#activities &	\#samples\\
         \hline
         \multirow{3}{1.7cm}{NTU RGB-AB}& train	&85&\multirow{3}{*}{94}&70952\\
         & gallery& \multirow{2}{*}{21}	&	&14192\\
         &probe&&&			3548\\
         \hline
         \multirow{3}{1.7cm}{PKU MMD-AB}& train	&53&\multirow{3}{*}{41}&13634\\
         & gallery& \multirow{2}{*}{13}	&	&2727\\
         &probe&&&			681\\
         \hline
         \multirow{3}{1.7cm}{Charades-AB}& train	&214&\multirow{3}{*}{157}&45111\\
         & gallery& \multirow{2}{*}{53}	&	&9022\\
         &probe&&&			2256\\
         \hline
         \multirow{3}{1.7cm}{MEVID}& train	&104&\multirow{3}{*}{1}&6338 (tracklets)\\
         & gallery& 52	&	&316 (tracklets)\\
         &probe&54&&			1438 (tracklets)\\
    \end{tabular}
    \label{tab:data_description}
\end{table}

\section{Future Work}
\label{sec:future}
While our method demonstrates strong performance in disentangling biometrics, non-biometrics, and motion features for activity-based person identification, there are areas for further exploration. The reliance on structured text supervision ensures effective feature separation, but future work could explore more flexible multimodal alignment techniques to further enhance robustness in unconstrained settings. Additionally, integrating a memory-modeling framework, could enhance identity tracking across much longer activity sequences, ensuring stability even under extreme motion variations or video length.

\end{document}